%% file: acl.tex
\pdfoutput=1

\documentclass[11pt]{article}

\usepackage{acl}

\definecolor{good}{RGB}{220,245,220}     %
\definecolor{partial}{RGB}{255,245,200}  %
\definecolor{bad}{RGB}{255,220,220}      %

\usepackage{amsmath}
\usepackage{amssymb}
\usepackage{amsthm}
\usepackage{thmtools}

\usepackage{times}
\usepackage{bm}
\usepackage{latexsym}
\usepackage{CJKutf8}
\usepackage{colortbl}
\usepackage[utf8]{inputenc}

\usepackage[utf8]{inputenc}

\usepackage{microtype}

\usepackage{inconsolata}

\usepackage{graphicx}
\usepackage{subcaption}
\usepackage{booktabs}
\usepackage{makecell}
\PassOptionsToPackage{table}{xcolor}
\usepackage[most]{tcolorbox}
\usepackage{thmtools}
\usepackage{soul}
\usepackage{hyperref}
\usepackage{soul}

\usepackage{tikz}

\usepackage{amssymb}
\usetikzlibrary{patterns}  %

\definecolor{White}{rgb}{1, 1, 1}
\definecolor{Periwinkle}{rgb}{0, 0, 0}

\newcommand{\ttt}[1]{\texttt{#1}}

\newcommand{\tlp}{\ensuremath{\mathit{TLP}}}
\newcommand{\tl}{\ensuremath{\mathit{TL}}}

\newcommand{\tbh}{translation barrier hypothesis}
\newcommand{\llama}{\texttt{llama-3.1}}
\newcommand{\aya}{\texttt{aya-23}}

\colorlet{LightGray}{White!98!Periwinkle}

\definecolor{mygreen}{RGB}{55,126,34}
\definecolor{myyellow}{RGB}{249,217,73}
\definecolor{gold}{RGB}{255,215,0}
\definecolor{darkorange}{RGB}{255, 140, 0}

\newtcolorbox{hypothesis}{
  enhanced jigsaw,
  colback=LightGray,
  drop shadow,
  boxrule=0.9pt,
  boxsep=0.1pt,
  left=4pt,
  right=4pt,
  top=4pt,
  bottom=4pt,
}

\title{\textit{The Translation Barrier Hypothesis:} \\  Multilingual Generation with Large Language Models Suffers from Implicit Translation Failure}

\author{Niyati Bafna\textsuperscript{1}, Tianjian Li\textsuperscript{1}, Kenton Murray\textsuperscript{1}, \\ \textbf{David R. Mortensen\textsuperscript{2}, David Yarowsky\textsuperscript{1}, Hale Sirin\textsuperscript{1}, and Daniel Khashabi\textsuperscript{1}} \\
         \textsuperscript{1}Johns Hopkins University, Center for Language and Speech Processing; \\ \textsuperscript{2}Language Technologies Institute, Carnegie Mellon University  \\
         \texttt{\{nbafna1,tli104\}@jhu.edu}
         }

\begin{document}

\maketitle
\begin{abstract}

Multilingual generation with large language models (LLMs) is often of poor quality for mid- to low-resource languages, but the causes for this are not well-understood.
We first demonstrate the existence of an implicit \emph{task-solving$\rightarrow$translation} pipeline for generation, whereby the model first solves the required task in a largely target-language-agnostic manner, and subsequently translates answer concepts into the intended target language.
We hypothesize that the failure of the translation stage, despite task-solving success, is an important culprit for the observed low quality of final outputs, and formalize this as the \emph{translation barrier hypothesis}.
We quantify the extent to which either stage in the pipeline is responsible for final failure for a word translation task across $108$ language pairs, and find that the translation barrier explains a dominant portion of error for a majority of language pairs, and is especially severe for low-resource target languages. 
Our results highlight an important bottleneck for end-to-end multilingual generation, relevant for future work seeking to improve multilinguality in LLMs.
\footnote{\href{https://github.com/JHU-CLSP/translation-barrier}{github.com/JHU-CLSP/translation-barrier}} 
\end{abstract}

\input{latex/1-introduction}

\input{latex/2-datasets}

\input{latex/3-method}

\input{latex/4-results_discussion}

\input{latex/6-related_works_shorter}

\input{latex/7-conclusion}

\section*{Limitations}

\paragraph{Scope}
This work only explores the \tbh{} for a word translation task.
It may be more difficult to assess task-solving success for more complex sequence-level tasks such as sentence-level machine translation, summarization, or open-ended instruction following in a principled manner using artifacts such as intermediate layer representations, since the nature of correct intermediate answer concepts could be less clear to ascertain.
There may also be additional complexities to take into account when extending our setup to these tasks: for example,  whether task-solving occurs ``globally'' over many parts or subtasks of a single input or ``locally'' for different fragments of the input, or the manner in which the target language syntax interacts with the ordering of the concepts in the task-solving stage.
Thus, our hypothesis may be difficult to verify for other tasks.

We clarify that our findings regarding the role played by the translation barrier may be dependent on the nature and complexity of the task, as well as the usage of strategies such as task- or language-specific finetuning or inference-time techniques such as in-context learning to boost multilingual performance.

While we do not claim that the statement of the hypothesis holds for all tasks and settings, we believe that the line of investigation in our work has important implications for future directions to improve the multilingual capabilities in LLMs, especially regarding the potential of end-to-end versus cascaded or modular approaches for the same.
Our goal is to invite discussion and further investigation of the translation barrier, as well as the most viable strategies for its mitigation.

\paragraph{Models and languages}

We conduct our experiments on two mainstream multilingual models; our findings should be verified on a range of different models.
Similar to previous work whose insights we build upon, we also only work with decoder-only models.
Our posited internal pipeline may look different for encoder-decoder models such as Aya-101 \citep{ustun2024aya}. 

Further, our findings are restricted to the $36$ target languages and $3$ source languages that we test on, and may differ for other languages and language families. 
In this work, we focus on mid- to low-resource range languages, including national languages like Swahili and Thai.
Future work may look at extending our setup and exploring the translation barrier for lower-resourced languages.

Finally, we use English as our prompt language regardless of the source language (i.e. the language of the term to be translated); this is as per work that says that this performs better \citep{dey2024better}.
Testing the effect of different prompt languages may add further nuance to our findings.

\section*{Ethics Statement}
We do not anticipate any negative consequences of this work, which is intended to diagnose poor multilingual capabilities in large language models.
We acknowledge that different language communities may have various and diverse needs from language technologies that may diverge from the points of focus of mainstream NLP, and not necessarily include LLM technologies \citep{bird-2020-decolonising}.
All the models we used are publicly available for research purposes: Aya-23-8B has a \ttt{CC-BY-NC} license, and the Llama family of models have a Llama 3.1 Community License Agreement, permitting research use. 
We will release all the artifacts created with the non-anonymized version of this paper. 
We used AI agents only for coding assistance (e.g. GitHub Copilot).

\section*{Acknowledgments}

We would like to thank Kaiser Sun and Krithika Ramesh for their helpful feedback. We would like to thank Jacob Wojnas, Varsini Sakthivadivel Ramasamy, and Suhas Sasetty, for providing annotations for the manual evaluation of the Polish, Tamil, and Telugu datasets respectively. This work is partially supported by the AI and Advanced Computing Institute at Schmidt Sciences.

\bibliography{acl}

\newpage
\appendix
\input{latex/appendix}

\end{document}

%% file: latex/1-introduction.tex
\section{Introduction}
\label{sec:intro}
Large language model (LLM) generation in mid-to-low-resource languages (LRLs) is of notoriously worse quality than that in high-resource languages (HRLs) \citep{robinson-etal-2023-chatgpt,hendy2023good,cahyawijaya-etal-2024-llms,jiao-etal-2023-chatgpt,languagebarrier2024shen}, and further suffers the problem of off-targetness, where an LM fails to output text in the intended language of generation \citep{zhang-etal-2020-improving, li-murray-2023-zero, marchisio-etal-2024-understanding}.
While ongoing work attempts to use explicit cascaded approaches, i.e. LLM generation into HRLs, followed by translation into the intended target, for multilingual generation \citep{shi2022language,liu2024translation}, or attempt to include support for more languages in LLMs for end-to-end generation, we still lack a systematic understanding of the reasons for and mechanisms of the above failures.

\begin{figure}[t]
  \centering
  \includegraphics[width=0.9\columnwidth]{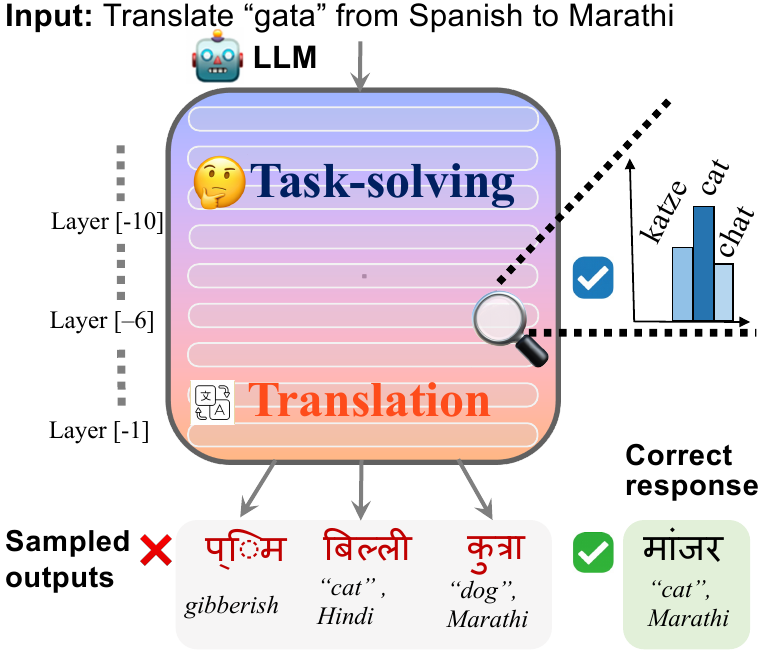}
  \caption{\emph{Task-solving} succeeds with the correct answer concept (i.e. \texttt{cat}) discovered in intermediate layers, expressed in various HRLs, but the model fails to realize or \emph{translate} the concept into the target LRL.}
  \label{fig:intro}

  \begin{hypothesis}
  \textbf{Translation Barrier Hypothesis:}
  \textit{Translation failure in LLMs accounts for a large proportion of poor quality final outputs for multilingual generation, where translation refers to model-internal transfer of answer concepts into the intended output language, given an implicit \emph{task-solving$\rightarrow$translation} cascade.}
  \end{hypothesis}
\end{figure}

Work from mechanistic interpretability shows that intermediate layer representations of multilingual LLMs are close to English representations even when prompted to generate in other languages, and that the models demonstrate language-specific behavior only in late layers of the model \citep{wendler2024llamas,foroutan-etal-2022-discovering,tang-etal-2024-language,kojima-etal-2024-multilingual}. 
Building upon this intuition, we posit a \emph{model-internal task-solving$\rightarrow$translation} cascade, with the middle layers of the LLM responsible for \emph{task-solving}, or discovering the required output concepts, in a target-language-agnostic manner, and the last few layers responsible for realizing those concepts in the target language, through \emph{translation}.

We present the \textbf{\tbh{}}, illustrated and stated in \autoref{fig:intro}, regarding the poor quality of multilingual generation.
This hypothesis has important consequences for end-to-end multilingual generation.
If translation failure, as opposed to task-solving failure, is the dominant culprit for low multilingual performance, then we can modularize or investigate architectural alternatives for the final LLM layers, or seek to induce better transfer in other ways, while still relying on powerful LLM task processing for end-to-end generation in LRLs. %
On the other hand, if task-solving capabilities themselves fail for LRL targets, explicit cascading methods become more viable options, calling for more investment in external specialized HRL-LRL MT systems. 

Our work quantifies the translation barrier on a word translation task, for two mainstream multilingual models, in order to demonstrate the extent of this issue. 
We choose word translation as a minimal generation task: in this simple case, \emph{task-solving} consists simply of realizing the semantics or the content of the input term, whereas \emph{translation} refers to generating the target language term for that concept. 
We use logit lens to retrieve model-internal hypotheses across different layers, assessing whether the model has arrived at the appropriate semantics for a particular input regardless of language; we also assess the on-target accuracy of the final outputs.
This allows us to demonstrate the existence of the posited pipeline.
We then test our hypothesis, by quantifying the extent to which either stage in the pipeline is responsible for final failure.

We make the following contributions:

\begin{itemize}
    \item We formalize the \textbf{translation barrier hypothesis}, highlighting a key mechanism of failure for multilingual generation with LLMs. 
    \item Working with a word translation task, we provide evidence for the existence of an implicit \emph{task-solving$\rightarrow$translation} pipeline for LLM generation over $36$ (target)  $\times \; 3$ (source) language pairs, and develop a framework within which we can measure and compare the contribution of \emph{task-solving} and \emph{translation} failure to total final failure.
    \item We characterize the languages in which task-solving occurs, going beyond English-centric previous work to show that intermediate layer representations are in fact multilingual to some extent and dominated by model-supported languages, with English foremost. 
    \item We find that \emph{translation} failure dominates ($>50\%$) final failure for $65\%$ and $78\%$ of all language pairs for \aya{} and \llama{} respectively, and that it especially forms the bottleneck for generation in low-resource or unsupported \emph{target} languages. On the other hand, we find that \emph{task-solving} is the bottleneck in the case of a low-resource \emph{source} language, and is also affected to some degree by the intended target language. 
    \item We provide case studies with a larger sized model and on a different task. These yield consistent findings with the above, inviting future work in understanding the translation barrier in different settings.
\end{itemize}

%% file: latex/2-datasets.tex
\section{Experimental Setup}
\label{sec:exp_setup}

\paragraph{Word translation dataset}
Given a word translation task between a source and target language, we are interested in judging the semantics of  intermediate layer outputs regardless of language.
We create a multiparallel dataset consisting of translations of words aligned across all languages of interest.
We take $400$ source words in English, split equally among nouns, verbs, adjectives, and adverbs,    %
and use the Google Translate API\footnote{\href{https://translate.google.com/}{translate.google.com}} to collect a set of translations per word, including synonyms, for all languages and for English itself. 
This allows us to construct a lexicon between any source and target language for the same concepts, as well as assess off-target accuracy in all other languages per concept.
We manually inspected the source words and lexicons for correctness and coverage (see \autoref{app:dataset_details} for more details).

\paragraph{Models}
We choose two performant mainstream models with a differing extents of multilinguality to conduct our experiments on: Aya-23-8B
(\aya) \citep{aryabumi2024aya} and Llama-3.1-8B-Instruct
(\llama) \citep{grattafiori2024llama3herdmodels}.
These are both 8-billion-parameter, decoder-only Transformer models, fine-tuned for instruction-following in $23$ and $8$ languages respectively. 
See \autoref{app:model_hyperparams} for generation hyperparameters and our prompting setup.

\paragraph{Languages}

By virtue of their data-hungry nature, mainstream LLMs only support a limited set of usually high- and mid-resource languages, but may be used and are claimed to demonstrate incident multilingualism in other languages \citep{blevins-zettlemoyer-2022-language}. 
We study $36$ target languages, including all languages supported by either model,\footnote{We count simplified and traditional Chinese as separate; thus we say that \aya{} supports $24$ languages. Thai is the only language supported by \llama{} but not \aya{}.} and a mix of $11$ additional mid- and low-resource languages unsupported by both models.
See \autoref{app:languages} for the full list of supported and unsupported languages and their resource level.
We use three different source languages against all the above targets: a high-resource supported language (Spanish), a mid-resource supported language (Hindi), and a mid-to-low resource unsupported language (Telugu). Note that ``source'' language refers to that of the translation task input; the task instruction is always provided in English.
We consider all the above $36$ languages as candidates for task-solving languages in intermediate model layers.

%% file: latex/3-method.tex
\DeclareRobustCommand{\patternbox}[3]{\tikz[baseline=-0.1ex]{\fill[#3] (0,0) rectangle (0.7em,1.3ex);
\fill[pattern=#1, pattern color=#2] (0,0) rectangle (0.7em,1.3ex);}}

\begin{figure*}[ht]
    \centering
    \includegraphics[scale=0.75]{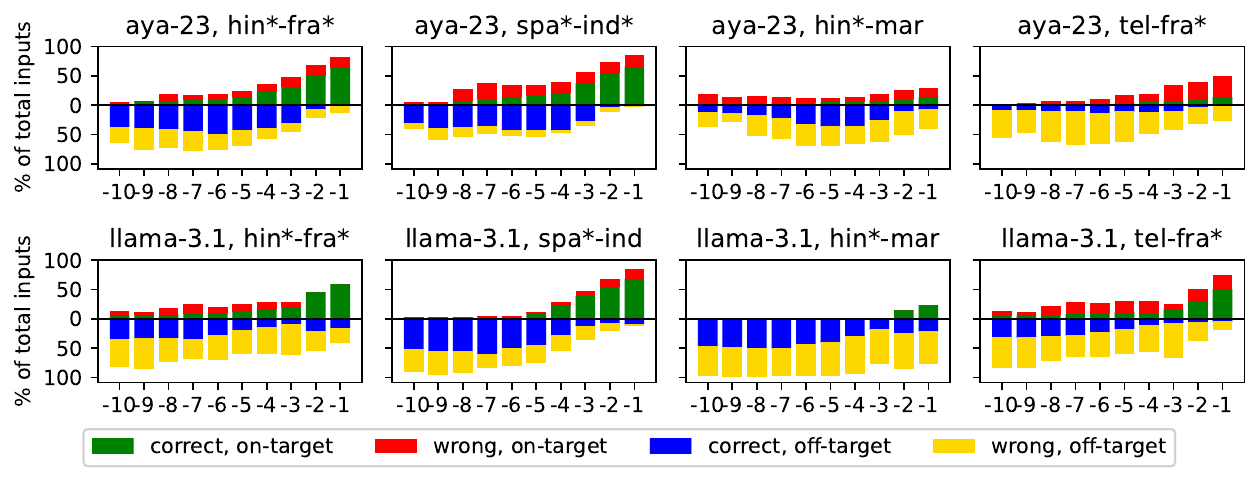}
    \caption{Each plot shows the percentage of \underline{on}-target correct (\patternbox{north east lines}{black}{mygreen}) and incorrect (\patternbox{north east lines}{red}{red}) outputs in the \textit{top half}, and \underline{off}-target correct (\patternbox{north west lines}{black}{blue}) and incorrect (\patternbox{north east lines}{myyellow}{myyellow}) outputs in the \textit{bottom half}. We show this for the last $10$ layers of \aya{} and \llama{}, for all outputs with a reliable LID tag. * supported language. We observe the \emph{task-solving} stage with initially high accurate but off-target outputs (\patternbox{north west lines}{black}{blue}), followed by the \emph{translation} stage, where the models transitions to on-target outputs (\patternbox{north east lines}{black}{mygreen}$+$\patternbox{north east lines}{red}{red}). We see that the translation stage is successful for French and Indonesian (HRLs) with high final on-target accuracy (\patternbox{north east lines}{black}{mygreen}), but fails for Marathi (LRL). For a low-resource source language like Telugu, \emph{task-solving} itself may fail, as with \aya{}, with low off-target accuracy (\patternbox{north west lines}{black}{blue}).
    }
    \label{fig:layerwise_evolution}
\end{figure*}

\section{Quantifying Task-solving and Translation failure}
\label{sec:method}

The \tbh{} presupposes the existence of a model-internal \emph{task-solving$\rightarrow$translation} pipeline (demonstrated and discussed in \autoref{sec:discussion}), with model success predicated on task-solving success followed by translation success.
We can study this hypothesis in the context of a given task by quantifying the extent to which translation failure is responsible for poor final outputs, as opposed to task-solving failure.

\paragraph{Obtaining intermediate layer outputs}
\label{sec:logit_lens}

We use logit lens \citep{nostalgebraist2020logitlens} to obtain text outputs from intermediate layers.
Let $x$ be an input prompt and $h_{i}(x)$ be the output of layer $i$. 
Logit lens applies the unembedding matrix $W_u$ and softmax to $h_i(x)$ to produce a probability distribution over the vocabulary at layer $i$:
\[
p_i(\cdot \mid x) = \operatorname{softmax}\left( W_u \cdot \operatorname{norm}\big( h_i(x) \big) \right)
\]

\noindent
where $\text{norm}$ refers to RMSNorm \cite{zhang-sennrich-neurips19} for \llama~and Layer Normalization \citep{ba2016layernormalization} for \aya.
We apply greedy decoding with this distribution to produce text output  $O(l_i)$ from the intermediate layer $l_i$.

Logit lens only operates at the token level; however, terms in our dataset may be multi-token. 
We iteratively append the most probable token of the final layer output (standard greedy decoding) to $x$ and apply a logit lens at every decoding step, enabling us to decode multi-token outputs from intermediate layers.

\paragraph{Defining translation loss}
Given a metric $M$ which measures the quality of final model outputs for our task, suppose that we can measure the quality of task-solving at intermediate layers with a comparable metric $M'$.
We then define \emph{translation loss} \tl{}
for a given source word $x$ and reference $\textbf{y}$,
\begin{equation}
TL(x) := \max_{i<L} [M'(O(l_i), \mathcal{\textbf{y}})] - M(O(l_L), \mathcal{\textbf{y}}), 
\label{eqn:tl}
\end{equation}
where $O(l_i)$ is the text output at model layer $l_i$, and $L$ is the total number of layers.
Aggregating instance level \tl{} over the dataset $\mathcal{D}$ yields the dataset level translation loss:
$$ TL(\mathcal{D}) = \sum_{x \in \mathcal{D}} TL(x) $$

The first term in \autoref{eqn:tl} measures task-solving success: we consider the task-solving performance at the best intermediate (non-final) layer for a given input.
The second term measures final success at output layer $l_L$, i.e. the success of the \emph{task-solving$\rightarrow$translation} pipeline. 
Thus \tl$(\mathcal{D})$, the difference between these, measures performance loss due to failed (internal) translation.

\paragraph{Defining $\bm{TLP}$ for word translation}
Our word translation dataset contains a set of equivalents $\{y^t\}$ per concept (source word $x$) per target language $t$.
The metric $M(y\hat{}, \textbf{y}=\{y^t\}) \in \{0, 1\}$ for measuring final accuracy is a binary exact match metric with any of the target language reference translations.
We consider the model to have task-solved successfully at a given layer if the layer output expresses the semantics of the input in \textit{any} non-source language. 
Thus, 
$$ M'(O(l_i),\textbf{y}) = \max_{t \in T, t \neq s} M (O(l_i), \textbf{y}=\{y^t\} )  $$

\noindent
where $T$ is the set of all languages and $s$ is the source language.
Given that the binary accuracy metric straightforwardly allows us to compute total failure as the number of source words with failed final translations $d_F$, we can now compute translation loss proportion (\tlp):
\begin{equation}
   TLP = TL(\mathcal{D}) / d_F 
   \label{eqn:tlp}
\end{equation}

Note that in this setup, the complement $1-$\tlp{} gives us task-solving failure as a fraction of total failure.
Thus, computing \tlp{} allows us to compare the proportion of blame that either stage of the pipeline bears for total final failure.

\paragraph{Layerwise analysis}
\label{sec:extracting}

For each layer output, we are interested in its task-solving accuracy, measured by the exact match metric $M'$, and its language.
We use our dataset language tag for the layer output in case of a match against our listed language equivalents.
For inaccurate outputs, we perform language ID on the layer output with the NLLB language ID model \citep{nllb2022}.\footnote{\href{https://huggingface.co/facebook/fasttext-language-identification}{huggingface.co/facebook/fasttext-language-identification/}}
We only retain the tag if it belongs to our set of target languages, which includes all languages supported by the model
i.e. plausible intermediate languages, and discard it if not. 
This is because intermediate layer output text may be gibberish or malformed, in which case we expect the language ID model to return a noisy random tag. 
See \autoref{app:lang_id_reliability} for a manual evaluation of the reliability of the resulting tags.
We track accuracy and language of output for the last ten layers of each model.\footnote{Preliminary experiments show that including earlier intermediate layers does not affect measurements of task-solving accuracy. Logit lens is also unreliable for early layers \citep{belrose2023eliciting}.}

%% file: latex/4-results_discussion.tex
\section{Results and Discussion}
\label{sec:discussion}

\subsection{Visualizing the pipeline}
\label{sec:visualizing_pipeline}

In \autoref{fig:layerwise_evolution}, we plot on-target and off-target accuracy for four language pairs over all outputs for which we found a reliable language tag. As targets, we include French and Indonesian as representative HRL targets, and Marathi as an example of an unsupported LRL target. 
We also include all three source languages. 
We find the patterns discussed below hold generally for all language pairs; see Appendix~\ref{app:evolution} for more examples.

We first observe the effect of the source language. 
While both models show high off-target accuracy with Hindi and Spanish as sources, \aya{} behaves differently with respect to Telugu: regardless of target, it yields low intermediate or final accuracy.
This indicates \aya{} simply fails to comprehend Telugu task inputs, resulting in failed task-solving even for HRL targets such as French.
This is less true of \llama{}, which shows similar patterns with Telugu as source as with Hindi and Spanish. 

We then look at the effect of the target language.
We consistently observe high off-targetness in middle layers, including for Spanish and other supported or high-resource languages, with considerable off-target accuracy, which subsequently decreases as the model pivots to the desired language.
For supported or HRL targets, there is a corresponding increase in on-target outputs and accuracy in later layers, indicating successful translation.
Note that while HRLs like Indonesian may not be supported by the LLM (\llama{}), we still see similar patterns for these languages as for supported HRLs, indicating that language resourcedness and presumably its presence in pre-training corpora play an important role in this regard.
On the other hand, for Marathi and other unsupported or low-resource languages, we simply observe a total drop in correct (on-target or off-target) intermediate outputs as the model moves away from off-target equivalents but also fails to translate correctly, resulting in poor final outputs.
\emph{This provides intuition for the translation barrier for LRL generation with LLMs.} 
We note that languages pairs with English as a target behave differently from the above, with on-targetness and accuracy emerging early and continuing until the final layer. 
This is not surprising, given the dominance of English in model training data.

The consistent nature of these plots, with off-target accuracy in middle layers (successfully or unsuccessfully) converted to on-target accuracy, can be interpreted as the model first ``task-solving'', or realizing correct answer concepts in some language(s), and then converting or translating those to the intended language. 
We find that $91.6\%$ and $92.6\%$ of all on-target final correct answers for all language pairs are also identified as correct at some intermediate layer in a \textit{non-target} language for \llama{} and \aya{} respectively (see Appendix~\ref{app:evolution} for language-wise breakdowns).
\emph{This, along with the above visualizations, shows evidence of a task-solving$\rightarrow$translation pipeline in the model.}
This also validates our technique of observing intermediate accuracy.

\subsection{Characterizing task-solving languages}
\label{sec:int_langs}

The above observations raise the question of what language(s) task-solving occurs in, referred to \emph{task-solving languages}, and whether these differ by language pair.
In order to investigate this, we first identify the task-solving layers of the model by locating the switch between the \emph{task-solving} and \emph{translation} stages of the generation pipeline, and then study the distribution of languages in the task-solving layers.

\paragraph{Layer of switch}
\begin{figure}
    \centering
    \includegraphics[scale=0.7]{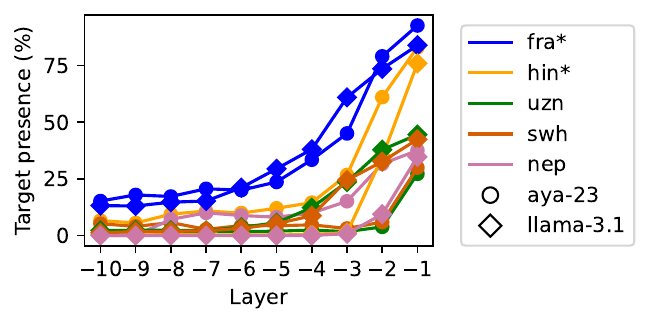}
    \caption{\% of target language presence among \textit{accurate} layer outputs for $5$ target languages, averaged over source language. * supported language. This stays low for middle layers, indicating that accurate answers are largely off-target, and increases in final layers, indicating \emph{translation} to the target language.
    }
    \label{fig:tgt_lang_presence}
\end{figure}

We use the proportion of the target language among all accurate layer outputs to track the switch between the two stages, since this will be high in the \emph{translation} stage, and plot this for five languages in \autoref{fig:tgt_lang_presence} for both models.\footnote{See Appendix~\ref{app:int_by_target} for complete plots. We also look at the proportion of target language outputs regardless of accuracy by layer; this results in similar trends.}

In general, both models show a significance increase in target language presence around the last four layers. 
We imagine that this is model- and task- dependent.
We also compute the specific layer with the maximum increase in target language presence from the previous layer, and find that this ranges $-1$ to $-4$ over different language pairs (see \autoref{app:layer_of_switch} for details). 
We found no correlation between the layer of switch and accuracy at the final layer over all language pairs.

\paragraph{Distribution over task-solving languages}

\begin{figure}[ht]
    \centering
    \includegraphics[width=0.85\linewidth]{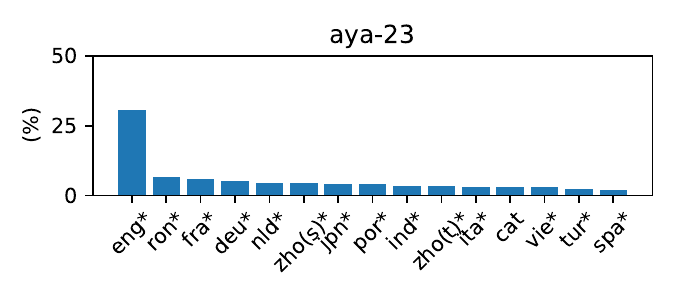}
    
    \vspace{-3mm}

    \includegraphics[width=0.85\linewidth]{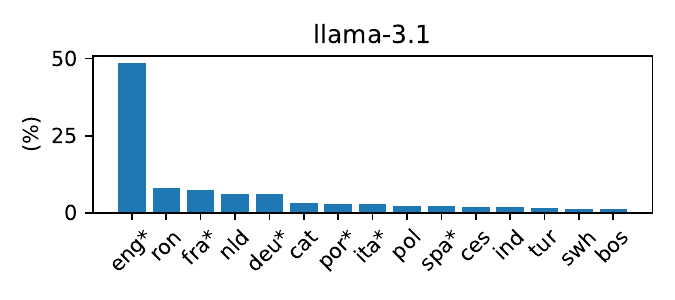}
    \vspace{-3mm}    
    \caption{Distribution over \emph{task-solving languages}, i.e. the languages of correct intermediate layer outputs, aggregated over all source-target language pairs for \aya{} (top) and \llama{} (bottom). We show the top $15$ task-solving languages, covering $85\%$ and $97\%$ of probability mass of the distribution for \aya{} and \llama{} respectively. * supported language.}
    \label{fig:int_lang}
\end{figure}

As per the above, we consider all layers up to the fourth-last layer as \emph{task-solving} layers,\footnote{This is a conservative heuristic approximation: as shown in \autoref{fig:tgt_lang_presence}, the switch between the stages is fairly fluid around the last few layers. Results look similar for surrounding choices of layer.} and look at the distribution over languages for all identified intermediate correct (off-target) answers in these layers.
We show the aggregate distribution over all language pairs in \autoref{fig:int_lang} for both models. 

We see that English dominates the intermediate representations for both models, although considerably more so for \llama. 
This is expected, given that \aya{} is much more multilingual than \llama. 
However, while previous work such as \citep{wendler2024llamas,etxaniz-etal-2024-multilingual,schut2025multilingual} focus exclusively on English as the intermediate representation language, we find that it only accounts for $30.7\%$ and $48.5\%$ of correct intermediate layer outputs for \aya{} and \llama{} respectively, with the rest of the distribution spread over other high-resource languages, with a tail of LRLs.\footnote{Note that we only compute this distribution over our considered $36$ languages. These include all model-supported languages: we observe that these constitute the bulk of the probability mass. Our experiments indicate that including other unsupported languages will largely only modify the long tail of the distribution.}
\emph{While this shows that multilingual LLMs are undeniably English-centric, it also demonstrates some extent of multilinguality, and raises questions about what information might be represented in other languages.}

We therefore also look at this distribution conditioned on target language (averaged over source languages), and find that it looks similar at middle layers, showing an increase of target language presence near final layers. (See Appendix~\ref{app:int_by_target} for details.)
\emph{Thus, we find that the task-solving stage of our demonstrated pipeline occurs in an English-dominant mix of supported and high-resource languages, and that this mix of languages is largely agnostic of the target language in middle layers.}

\begin{figure*}[!ht]
    \centering
    \hspace*{-10mm}\includegraphics[scale=0.7]{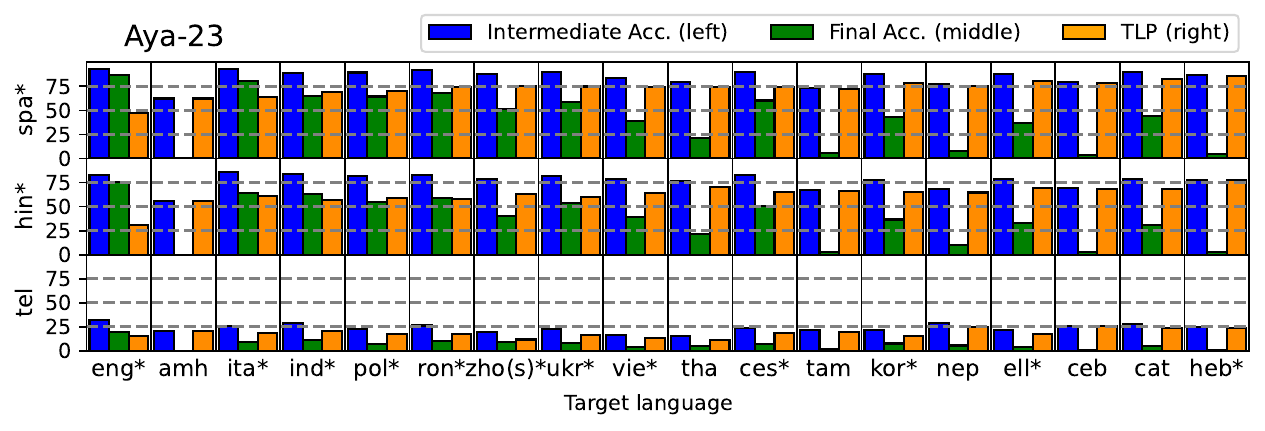}
    \hspace*{-10mm}\includegraphics[scale=0.7]{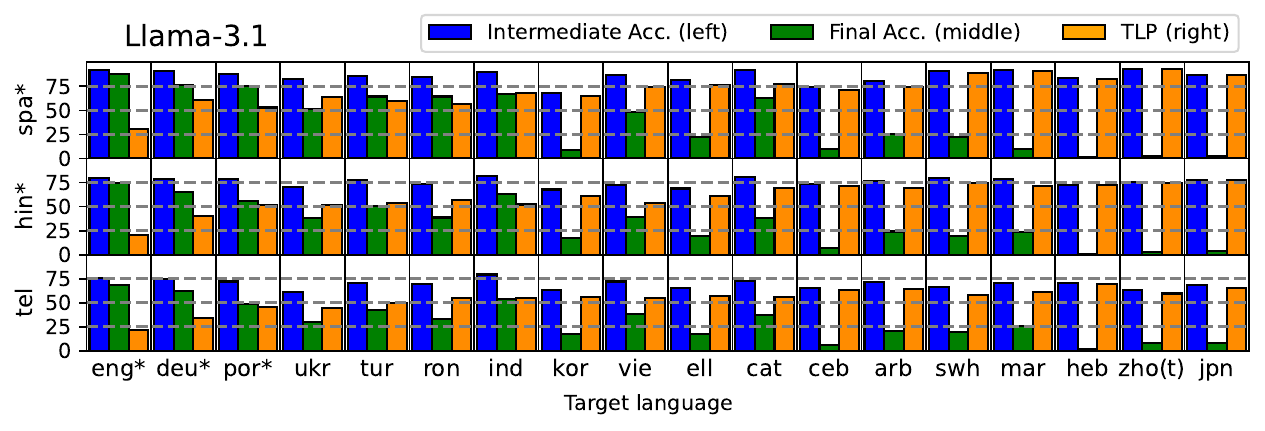}
    \caption{Intermediate accuracy (\patternbox{north east lines}{blue}{blue}), final accuracy (\patternbox{north east lines}{mygreen}{mygreen}), and \tlp{} (\patternbox{north east lines}{darkorange}{darkorange}, \autoref{eqn:tlp}) for \aya{} and \llama{}, sorted in ascending order of mean \tlp{} over source language for $18$ target languages, selected to cover the range of mean \tlp{}. *: supported language. While intermediate accuracy (\patternbox{north east lines}{blue}{blue}) is high even for LRLs (Cebuano-ceb, Nepali-nep), final accuracy (\patternbox{north east lines}{mygreen}{mygreen}) is high for supported HRLs (Portuguese-por, German-deu), but drops considerably for LRLs. \tlp{} (\patternbox{north east lines}{darkorange}{darkorange}) is high for most target languages, and especially for low-resource languages. See expanded figure including all target languages in \autoref{sec:all_results}.
    }
    \label{fig:main_results}
\end{figure*}

\subsection{Quantifying the translation barrier}
\label{sec:quant_tb}

\autoref{fig:layerwise_evolution} suggests that the model is often able to solve the task, but fails to generate / translate correctly into LRLs.
We quantify this effect by computing: (a) task-solving success or intermediate accuracy (the first term in \autoref{eqn:tl}): the percentage of task outputs that were correct at \textit{some} intermediate layer in \textit{some} language (b) final success (the second term in \autoref{eqn:tl}): on-target accuracy at the final layer, and (c) \tlp, as per \autoref{eqn:tlp}: the proportion of total failure that can be attributed to translation failure.

First, we look at these three quantities averaged over all target languages for each source language in \autoref{tab:summary_statistics}.
We see that intermediate accuracy is much higher than final accuracy in all cases, with much less variance, indicating that the former is stabler and less dependent on the target language than the latter.
This ties in with our understanding of the relatively target-language agnostic task-solving step, followed by a target-language dependent translation step.
\emph{Overall, translation loss accounts for a high percentage of total failure (more than $50\%$ in all cases except one). this lends evidence to the \tbh{}.}

\input{latex/figures/table_summary_statistics}

We also see that final and intermediate accuracy are roughly similar for both models, with the exception of for Telugu, for which \aya{} is much worse. 
This is perhaps surprising given than \aya{} is more multilingual than \llama, supporting $24$ as opposed to just $8$ of our $36$ target languages.

\paragraph{Role of translation failure}
Next, we look at intermediate accuracy, final accuracy, and \tlp{} by language pair and model.
We show these results for $18$ target languages against each source language for both models in \autoref{fig:main_results}, with the complete results for all $36$ target languages in \autoref{sec:all_results}. 
We find that translation failure dominates total failure ($>50\%$) for $65\%$ and $78\%$ of all language pairs for and \aya{} and \llama{} respectively. 
This provides direct evidence for the \tbh{} for individual language pairs.

We further find distinct patterns for high-resource versus low-resource target languages.
We observe that intermediate accuracy stays high even for low-resource unsupported target languages (like Nepali and Hebrew) for Hindi and Spanish as sources: in line with our previous observations, this indicates that the model is able to correctly task-solve even for low-resource targets.
In contrast, and as expected, we observe that final accuracy is high for supported target HRLs (like German, Italian, and Portuguese, for both models) but drops drastically for low-resource target languages (like Tamil and Swahili), some of which are supported, such as for Greek with \aya. 
Script and language family may naturally also contribute to this performance drop (e.g. \llama{} performs worse for Japanese and Arabic than for Romanian).
High intermediate and final accuracy as in the first case (HRLs) means that \tlp{} is relatively low: a larger part of total failure comes from task-solving errors, and the model is more capable of generating successfully in the case of task-solving success.
However, high intermediate accuracy and low final accuracy as in the second case (LRLs) indicates that translation loss accounts for a much bigger percentage of total error, going up to $82.3\%$ for \ttt{spa-cat} with \aya{}, and $91.1\%$ for \ttt{spa-mar} with \llama{}.

Leaving aside language pairs with English as a target, or with Telugu as a source, translation failure dominates total failure (\tlp $>50\%$) for $100\%$ and $90\%$ of language pairs for \aya{} and \llama.
English is a special case: given that it dominates the internal representations of these models, generation into English may have a minimal or absent translation stage, meaning that task-solving success largely determines final success.
We discuss the case of Telugu as a source below.

\emph{The above observations indicate that translation loss plays an outsized role in final failure even for supported target languages, and an even bigger role for low-resource or unsupported languages. These results validate the \tbh{} for the word translation task.}

\paragraph{Evidence for entanglement in the task-solving stage}
The above discussion generally supports our understanding of target-language-agnostic \emph{task-solving} stage followed by a target-language-specific \emph{translation} stage. 
However, we note the caveat that there is a clear disparity between intermediate task-solving accuracy across target languages, with the largest difference of $32.9$ percentage points between \ttt{spa-eng} and \ttt{spa-amh} with \aya, as well as differences between supported languages ($12$ percentage point difference between \ttt{hin-eng} and \ttt{hin-jpn} with \aya).
\textit{This indicates some level of entanglement between task-solving success and language of generation, meaning that the intended language of generation affects the model's ability to task-solve to some extent. 
}This hints that even in the hypothetical case of no translation loss, end-to-end processing for low-resource or unsupported languages would still fall behind that for HRLs due to imperfect transfer of task-solving abilities to LRLs. 
This finding calls for an evaluation of the future of end-to-end multilingual generation with LLMs on a larger scale of languages, with one alternative being to invest in better explicit output-side LLM+MT cascades.

\paragraph{Impact of source language}

We see that for both models, mean intermediate and final accuracy as well as mean \tlp{} are highest for Spanish, then Hindi, then Telugu (\autoref{tab:summary_statistics}).
It is unsurprising that intermediate accuracy and therefore final accuracy depend on the source language: task-solving success is naturally conditioned on comprehension of the task inputs.
\emph{These findings justify exploring an input-side MT+LLM cascade for different language task inputs even for supported languages.}

This also explains why \tlp{} is lowest for Telugu: 
with low intermediate accuracy, translation loss can only occur on very few inputs as a percentage of the total failure, which is dominated by failed task-solving cases.
For example, since task-solving fails for $74\%$ of total inputs for Telugu-Cebuano with \aya{}, task-solving failure dominates total failure even though the translation stage also fails on nearly all correct intermediate outputs.

\subsection{Case study: Effect of scale}
\label{sec:scaling}

\begin{figure}
    \centering
    \hspace*{-4mm}\includegraphics[scale=0.8]{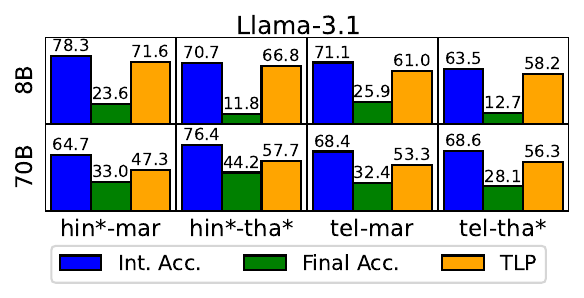}
    \caption{Intermediate accuracy (\patternbox{north east lines}{blue}{blue}), final accuracy (\patternbox{north east lines}{mygreen}{mygreen}), and \tlp{} (\patternbox{north east lines}{darkorange}{darkorange}, \autoref{eqn:tlp}) for \ttt{llama-3.1-8B} (top) and \ttt{llama-3.1-70B} (bottom). *: supported language. Final accuracy (\patternbox{north east lines}{mygreen}{mygreen}) is higher for the latter. \tlp{} (\patternbox{north east lines}{darkorange}{darkorange}) is lower but remains high in absolute terms.
    }
    \label{fig:scaling}
\end{figure}

We repeat our experiments for $4$ language pairs on Llama-3.1-70B-Instruct and show the effect of using a larger model on \tlp{} in \autoref{fig:scaling}. 
We find that the $70B$ model shows similar intermediate accuracy on average as the $8B$ model, but consistently higher final accuracy, or capability to produce the target language, resulting in lower \tlp{}.
However, absolute \tlp{} values remain high. 
These preliminary results indicate that the translation barrier may continue to be a problem at scale. 
\begin{figure}
    \centering
    \hspace*{-4mm}\includegraphics[scale=0.8]{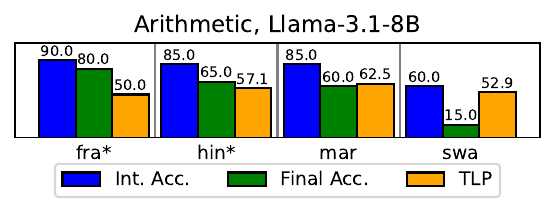}
    \caption{Intermediate accuracy (\patternbox{north east lines}{blue}{blue}), final accuracy (\patternbox{north east lines}{mygreen}{mygreen}), and \tlp{} (\patternbox{north east lines}{darkorange}{darkorange}, \autoref{eqn:tlp}) for \ttt{llama-3.1-8B} for an arithmetic task. Intermediate acc. stays high, but final accuracy is lower for LRLs, leading to high $TLP$.
    }
    \label{fig:arithmetic}
\end{figure}

\subsection{Case study: Arithmetic Task}

We conduct a small case study to investigate the translation barrier on an arithmetic task.
We used $20$ word arithmetic English questions (such as ``Eight times two equals?''), with numeral word answers, and prompted the model to respond in a target language, for $4$ variously resourced languages.
We manually examine the outputs and observe the task-solving$\rightarrow$translation pipeline, with the correct solved answers in HRLs in intermediate layers, with a switch to the target language in late layers. (See \autoref{app:arithmetic} for examples.)
We report intermediate and final accuracy, and $TLP$ in \autoref{fig:arithmetic}.
We observe high $TLP$ especially for lower-resourced languages; these results are consistent with our findings on word translation. 
We invite work in better understanding the translation barrier for other tasks and model sizes.

%% file: latex/figures/table_summary_statistics.tex
\begin{table}[ht]
\centering
\small
\setlength{\tabcolsep}{3pt}
\begin{tabular}{p{0.8cm}p{0.6cm}rrr}
\toprule
 & \textbf{Src.} & \textbf{Final Acc.} & \textbf{Int. Acc.} & \textbf{TLP} \\
\midrule
\ttt{aya}    & \ttt{spa*} & 42.0 $\pm$ 28.1 & 84.6 $\pm$ 7.2 & 71.6 $\pm$ 7.3 \\
\ttt{-23}    & \ttt{hin*} & 37.4 $\pm$ 24.3 & 76.1 $\pm$ 7.2 & 59.6 $\pm$ 8.4 \\
        & \ttt{tel}  & 9.5  $\pm$ 13.0 & 25.3 $\pm$ 11.3 & 17.5 $\pm$ 4.0 \\
        & \textbf{Avg} & \textbf{29.6} & \textbf{62.0} & \textbf{49.6} \\
\midrule
\ttt{llama}  & \ttt{spa*} & 39.8 $\pm$ 29.4 & 84.2 $\pm$ 9.8 & 68.0 $\pm$ 18.3 \\
\ttt{-3.1} & \ttt{hin*} & 33.7 $\pm$ 23.8 & 75.2 $\pm$ 7.8 & 58.2 $\pm$ 15.3 \\
        & \ttt{tel}  & 31.6 $\pm$ 21.1 & 69.4 $\pm$ 7.8 & 52.1 $\pm$ 12.6 \\
        & \textbf{Avg.} & \textbf{35.0} & \textbf{76.3} & \textbf{59.4} \\
\bottomrule
\end{tabular}
\caption{Mean and std. dev. of final accuracy (``final acc.''), intermediate  accuracy (``int. acc.''), and \tlp{} (all in \%), over all target languages for each source language, for both models. * supported language.
We see that intermediate accuracy is higher than final accuracy, with much less variance, indicating that the former is stabler and less dependent on the target language than the latter.
}
\label{tab:summary_statistics}
\end{table}

%% file: latex/6-related_works_shorter.tex
\section{Related Work}

While multilingual LLMs are trained to perform well for several languages \citep{lin-etal-2022-shot, workshop2023bloom176bparameteropenaccessmultilingual, aryabumi2024aya}, their pre-training data is dominated by English \citep{xue-etal-2021-mt5, chung2023unimaxfairereffectivelanguage, li-etal-2025-upsample}. 
\citet{etxaniz-etal-2024-multilingual} and \citet{zhao2024how} show that explicit cascading with MT often outperforms end-to-end multilingual inference.
Further, works such as \citet{wendler2024llamas} and \citep{schut2025multilingual} apply logit lens to intermediate layers and observe that multilingual LLMs ``think'' in English, or that intermediate representation spaces of these models lie close to English.
This is reinforced by works studying neuron activations for multilingual processing \citep{foroutan-etal-2022-discovering,tang-etal-2024-language,kojima-etal-2024-multilingual}, which find language-specific neurons only in later model layers.
Concurrent works to ours such as \citet{lu2025paths} and \citet{wang2025lost} demonstrate that late layer translation failure affects cross-lingual consistency and accuracy for factuality recall. 
Our work contributes to this discourse by formalizing and quantifying the translation barrier for a core generation task over a range of languages, with a focus on its impact on low-resource languages.

%% file: latex/7-conclusion.tex
\section{Conclusion}

We formalize the \textbf{\tbh} for poor quality multilingual generation with LLMs.
Working with a minimal generation task for $108$ language pairs, we visualize a \emph{task-solving$\rightarrow$translation} pipeline for multilingual generation, demonstrating its failure for low-resource languages.
We then quantify the role of the translation barrier as an important culprit for low final performance, discussing its relevance for differently resourced source and target languages.
Our findings contribute insights for future strategies in multilingual generation with LLMs.

%% file: latex/appendix.tex
\newpage
\onecolumn
\section{Languages}
\label{app:languages}
See \autoref{table:languages} for the list of languages that we studied.
\input{latex/tables/language_table}

\newpage

\section{Dataset Details}
\label{app:dataset_details}

In order to curate the source concepts that we translate to every language, we prompted ChatGPT to give us $200$ instances each of English nouns, verbs, adjectives and adverbs (totalling $800$ words). 
We then filtered out any words that did not appear in the English NLTK WordNet \cite{bird-loper-2004-nltk,miller-1994-wordnet}, and manually vetted the remaining words to retain ``general'' concepts (i.e. excluding any proper nouns as well as highly specific cultural terms). 
We also checked for offensive words and did not encounter any.
Finally, we sample $100$ words per part-of-speech of the filtered words, and conduct all our experiments on the resulting $400$ concepts. 

We collect translations, including synonyms and alternate word senses, for these $400$ concepts for each of the other $35$ languages studied using Google Translate as mentioned in \autoref{sec:exp_setup}.
For average number of equivalent words per source word per language, see \autoref{table:lexica_equivalents}.
We would like to note that the above method yielded only 1 equivalent per source word in some languages, like Chinese and Amharic and that this may have affected our results for these languages.
\input{latex/tables/lexica_equivalents}

\paragraph{Manual evaluation}

To evaluate the expanded alternative translation outputs, we sampled 100 words from two languages (Turkish and Polish) and manually evaluated the alternative translation groups for each source word for plausibility (on a scale of 1 to 4, 1 = ``none of the words are plausible'' and 4 = ``all of the words are plausible'') and for coverage (on a scale of 1 to 4, 1 = ``none of the word senses are covered'' and 4 = ``all of the word senses are covered''). We report 3.9 and 3.4 for coverage, and 3.8 and 3.7 for plausibility, for Polish and Turkish respectively.

We further asked proficient speakers of Hindi, Marathi, Uzbek, Telugu, and Tamil, to go through $100$ entries of our dataset, and report the number of entries with incorrect listed translations. Note that the last four languages are among the lowest-resourced in study. We report the resulting error rates in \autoref{app:error_rates}.

The consent of all annotators was obtained regarding the usage of the collected data for accuracy and coverage reporting of the lexica. 
We did not provide any monetary compensation for this annotation. 

\begin{table*}[t]
\centering
\begin{tabular}{l c}
\toprule
\textbf{Language} & \textbf{Error \%} \\
\midrule
Hindi   & 2\% \\
Marathi & 7\% \\
Uzbek   & 0\% \\
Tamil   & 1\% \\
Telugu  & 2\% \\
\bottomrule
\end{tabular}
\caption{Error rates in our dataset for 5 languages.}
\label{app:error_rates}
\end{table*}

\newpage
\section{Model hyperparameters and prompting setup}
\label{app:model_hyperparams}

\paragraph{Hyperparameters and compute}

We use greedy decoding with a temperature of $0$ and the official chat template of \aya{} and \llama, respectively.
We run inference with logit lens on NVIDIA H100 NVL Tensor Core GPUs.
This takes about $30$ minutes to process the test set of size $400$ with batch size $8$ on a single GPU, for each of our $108$ language pairs for each of \aya{} and \llama. We downloaded \aya{} and \llama{} from Huggingface Transformers \citep{wolf-etal-2020-transformers}.

We additionally run experiments on $4$ language pairs with \ttt{llama-3.1-70B}. 
This takes $45$ minutes per language pair on $4$ GPUs with a batch size of $8$.

This does not include exploratory experiments and debugging. 

\paragraph{Prompting}

Given that LLM performance is sensitive to the wording of the prompt \citep{anagnostidis2024susceptiblellmsinfluenceprompts}, we try a few different prompts for the word translation task. 
We find that final performance differs slightly with different wordings of the instruction as well as the manner in which the target language is specified.

Here are examples of prompts we explored:
\begin{enumerate}
    \item \ttt{Translate the following word into English:}
    
    \ttt{gata}
    \item \ttt{Translate the following word from spa\_Latn to mar\_Deva. Respond with a single word.} 
    
    \ttt{Word: gata}
    
    \ttt{Translation: }

    \item \ttt{Give a one-word translation of the following word from Spanish to Marathi.} 
    
    \ttt{Word: gata}
    
    \ttt{Translation: }

    \item \ttt{Translate the following word from Spanish to Marathi. Respond with a single word.} 
    
    \ttt{Word: gata}
    
    \ttt{Translation: }

\end{enumerate}

We use prompt (4) for all our experiments, as performing well for both our models.

\newpage
\section{Examples of inputs and outputs}
\label{app:input_and_outputs}

See \autoref{tab:int_layer_examples} for examples of intermediate layer outputs for some inputs.

\begin{table*}[!ht]
    \centering
    \includegraphics[width=0.7\linewidth]{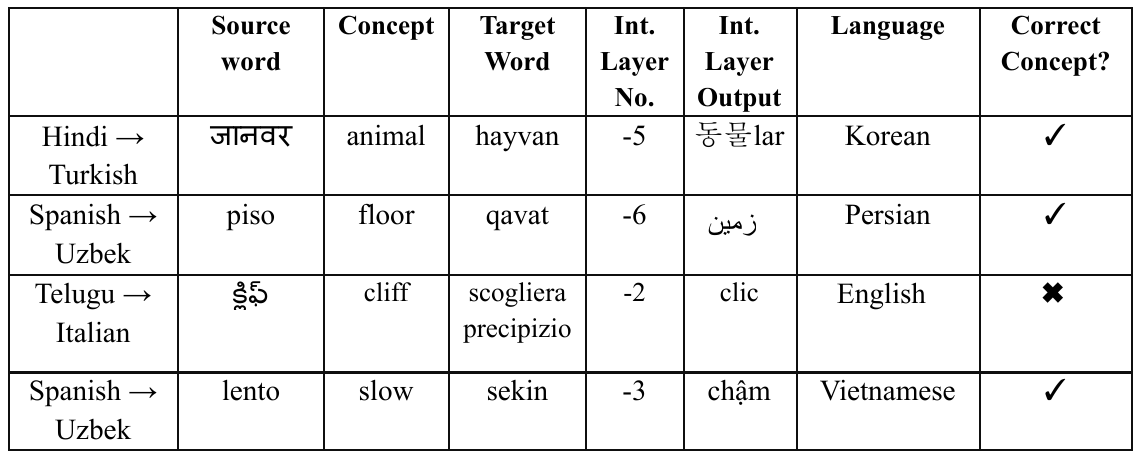}
    \caption{Intermediate layer output examples that are in a language other than the input or the target language. The prompt with the source word is formatted as mentioned in \autoref{app:model_hyperparams}.
}
    \label{tab:int_layer_examples}
\end{table*}

\newpage
\section{Reliability of the LangID}
\label{app:lang_id_reliability}

Recent work shows that neural language identification models can be performant even for mixed-script and noisy  multilingual data \cite{sirin-etal-2024-detecting}. To evaluate the language ID performance of the NLLB model \cite{nllb2022} over intermediate outputs, we sampled $100$ source-target word pairs from two different language pairs, Hindi - Turkish and Spanish - German. 
Note that we discard labels that do not correspond to one of our $36$ target languages as per \autoref{sec:method}, and only report on-targetness or off-targetness in \autoref{fig:layerwise_evolution} for inputs for which we can find a reliable label.

We evaluated the resulting language identification labels for two different layer outputs for each word pair.
We found that all labeled samples had outputs that were recognizably in some language. 
We marked the LID as correct if it accurately identified the language of any of the words in the output, given that the outputs could be language-mixed. 
We found that the model was $80\%$ accurate over intermediate layer outputs.

\newpage
\section{Evolution through layers for all languages}
\label{app:evolution}

\paragraph{Demonstrating abstract task-solving}

\begin{figure*}
    \centering
    \includegraphics[scale=0.4]{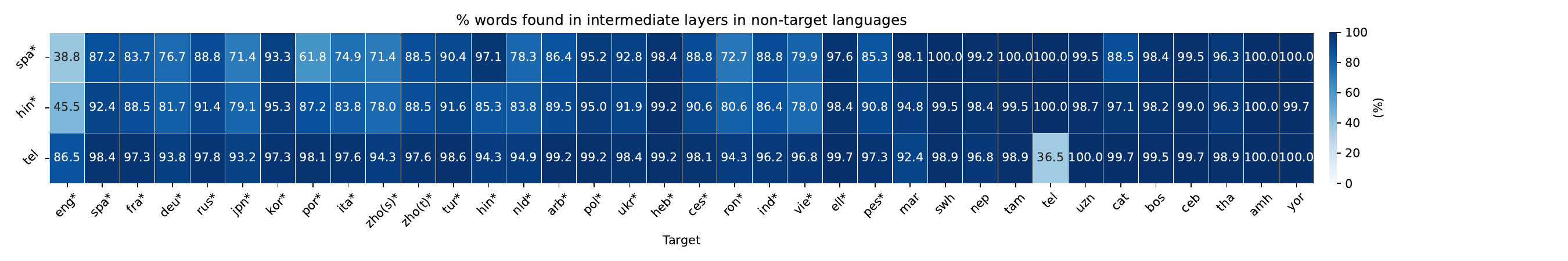}
    \includegraphics[scale=0.4]{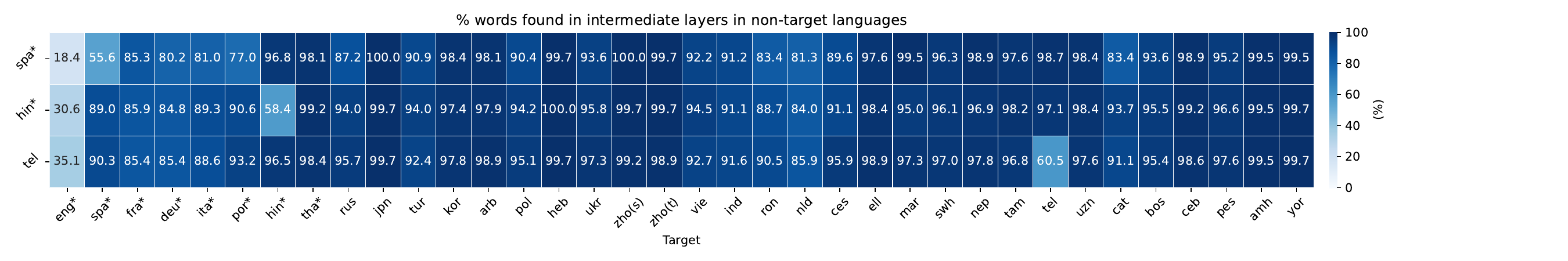}
    \caption{This shows the percentage of task inputs that had final correct answers for which we also correct intermediate layer outputs in a different (non-target) language for \aya{} (top) and \llama{} (bottom). This demonstrates that intermediate layers perform task-solving at some layer of abstraction.}
    \label{fig:correct_at_intermediate}
\end{figure*}

We show the percentage of final answers that were identified as correct in some intermediate layer, in a \textit{different language from the target}, in \autoref{fig:correct_at_intermediate} for both models.
In general, we see that this percentage is high, although less so for HRL supported languages. 
This, in conjunction with the shape of the distribution over intermediate languages that are largely target-language agnostic as discussed in \autoref{sec:discussion} and Appendix~\ref{app:int_by_target}, demonstrates that intermediate layers perform task-solving at some layer of abstraction.

\paragraph{Layer-wise accuracy for other language pairs}

\begin{figure*}[htbp]
    \centering
    \begin{minipage}{0.45\textwidth}
        \includegraphics[width=\linewidth]{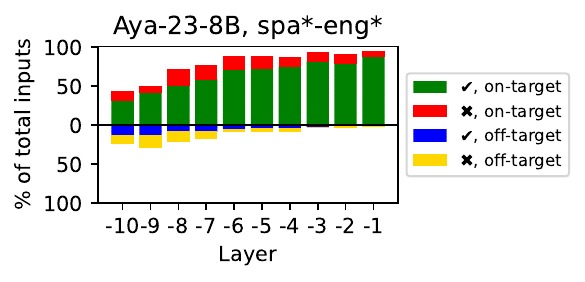}\\[1ex]
        \includegraphics[width=\linewidth]{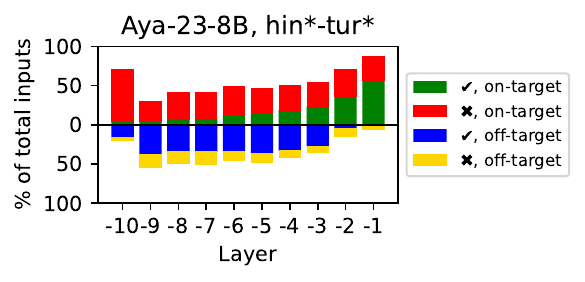}\\[1ex]
        \includegraphics[width=\linewidth]{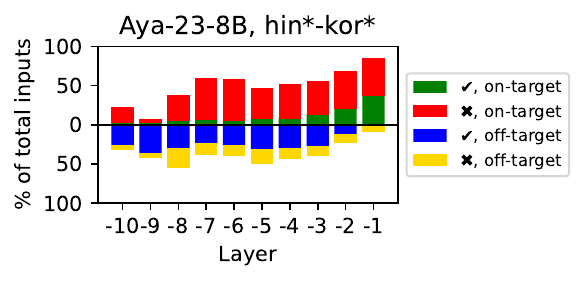}\\[1ex]
        \includegraphics[width=\linewidth]{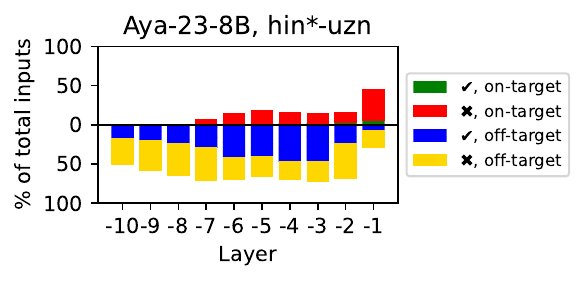}\\[1ex]
        \includegraphics[width=\linewidth]{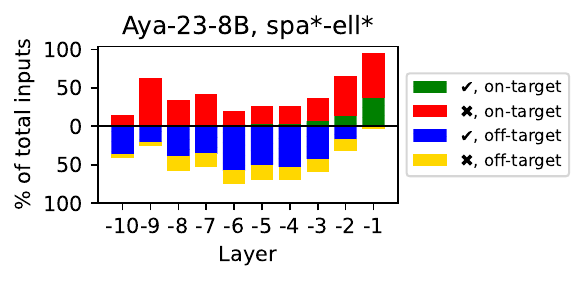}\\[1ex]
        \includegraphics[width=\linewidth]{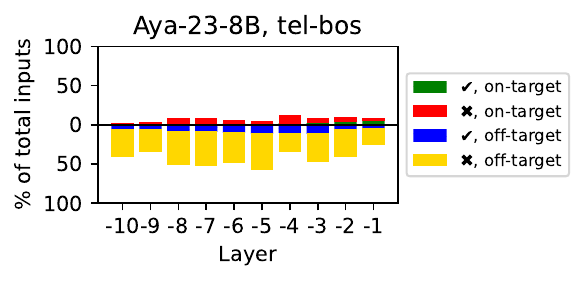}
    \end{minipage}
    \hfill
    \begin{minipage}{0.45\textwidth}
        \includegraphics[width=\linewidth]{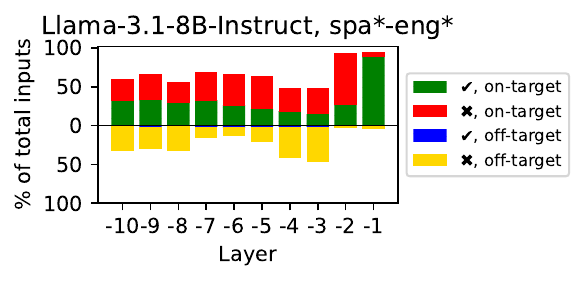}\\[1ex]
        \includegraphics[width=\linewidth]{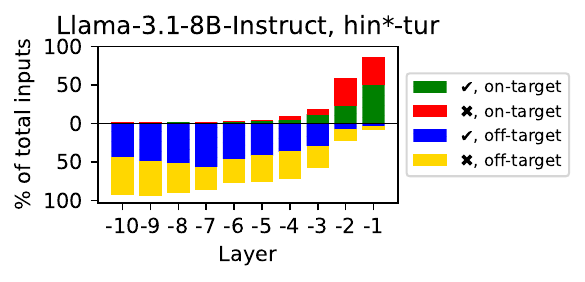}\\[1ex]
        \includegraphics[width=\linewidth]{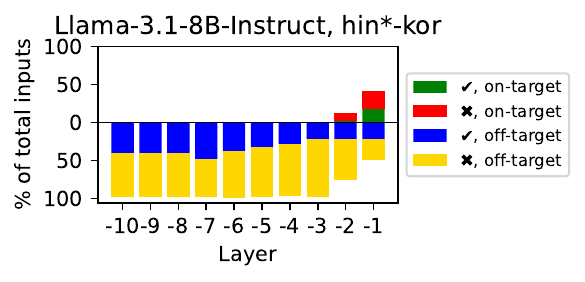}\\[1ex]
        \includegraphics[width=\linewidth]{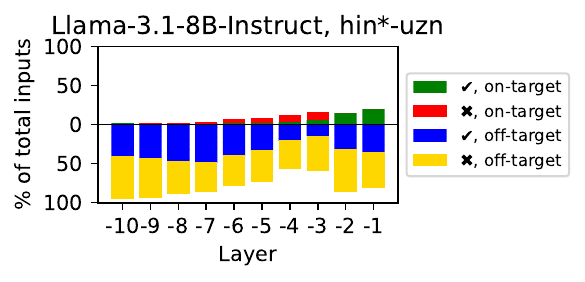}\\[1ex]
        \includegraphics[width=\linewidth]{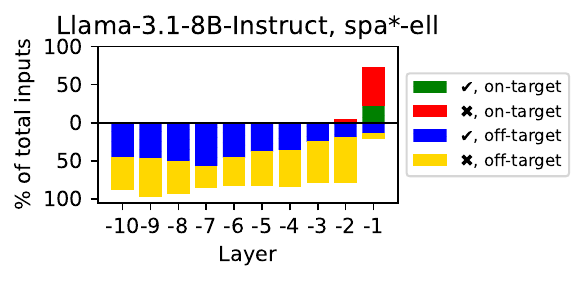}\\[1ex]
        \includegraphics[width=\linewidth]{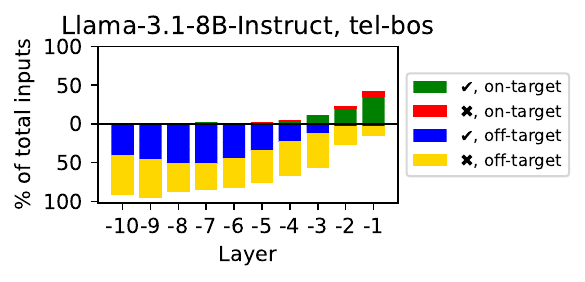}
    \end{minipage}
    \caption{Layerwise analysis plots for five language pairs: \aya{} (left) vs. \llama{} (right) (extension of \autoref{fig:layerwise_evolution}}
    \label{fig:evolution-grid}
\end{figure*}

We choose five representative language pairs and show the evolution of on-target and off-target accuracy through various layers for each of them for \aya{} and \llama{} in \autoref{fig:evolution-grid}. 
These cover all three source languages, supported HRL and MRL languages, as well as unsupported languages.

\newpage
\section{Layer of switch for individual language pairs}
\label{app:layer_of_switch}

\begin{figure*}
    \centering
    \includegraphics[width=\linewidth]{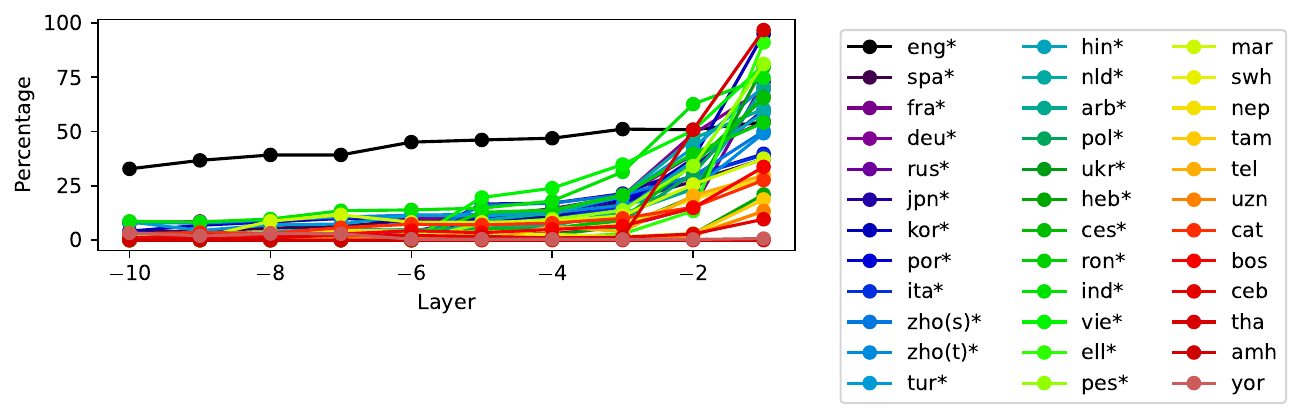}
    \includegraphics[width=\linewidth]{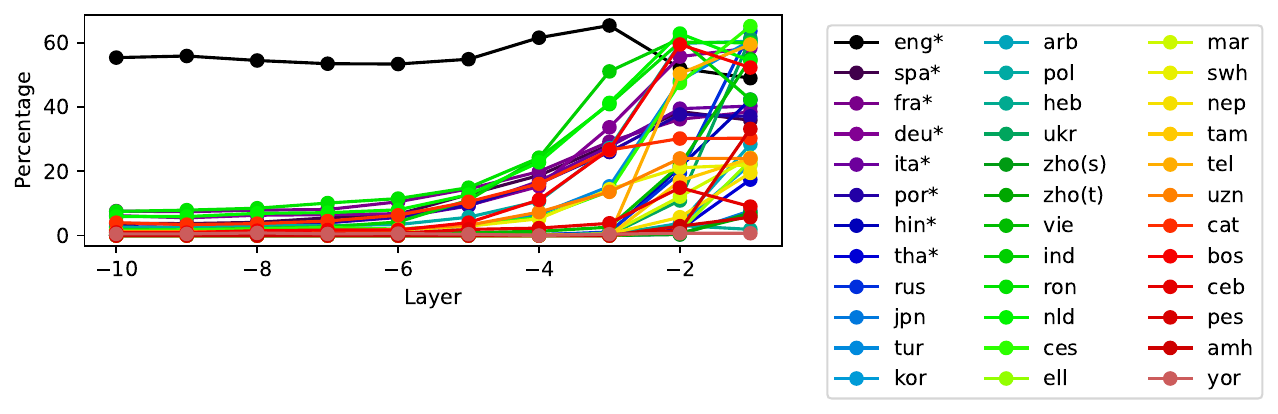}
    \caption{Target language presence by proportion of all languages for correct intermediate answers for \aya{} (top) and \llama{} (bottom) (complete version of \autoref{fig:tgt_lang_presence} for all target languages).}
    \label{fig:tgt_lang_by_layer}
\end{figure*}

See \autoref{fig:tgt_lang_by_layer} for a complete version of \autoref{fig:tgt_lang_presence} for all target languages. 

\begin{figure*}
    \centering
    \includegraphics[width=\linewidth]{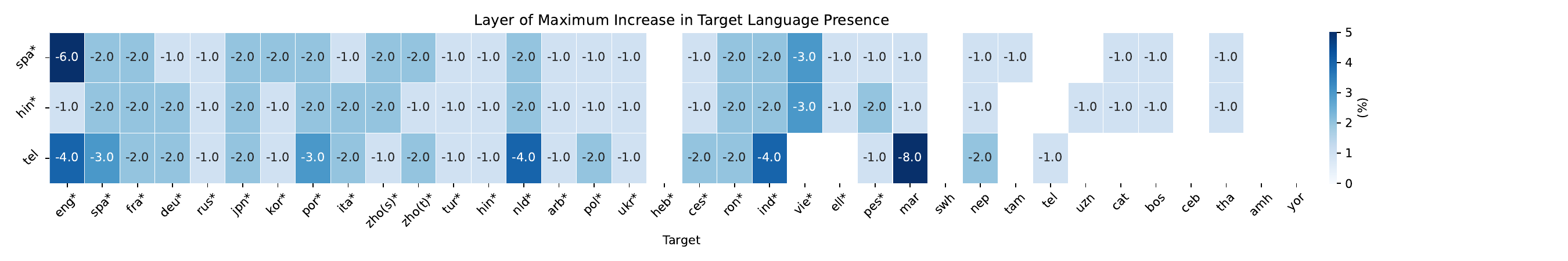}
    \includegraphics[width=\linewidth]{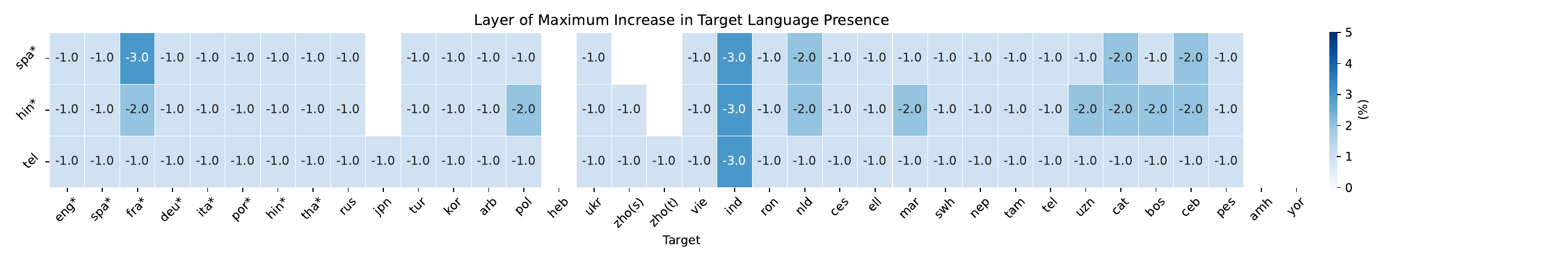}
    \caption{The layer with the maximum increase in target language presence for accurate outputs for \aya{} (top) and \llama{} (bottom). Only shown for language pairs that had $>5\%$ final accuracy.}
    \label{fig:layer_of_switch}
\end{figure*}

In \autoref{fig:layer_of_switch}, we show the layer with the maximum increase in target language presence for accurate outputs, in order to identify a particular layer of switch between the \emph{task-solving} and \emph{translation} stages.
We exclude language pairs that have negligible final accuracy ($<5\%$), since these effectively have no translation stage.
We see the layer of switch is on or after the fourth-last layer for \aya{} and on or after the third-last layer for \llama, with \aya{} often showing earlier switches than \llama. 

\newpage
\section{Distribution of ``task-solving'' languages by target language}

\label{app:int_by_target}

\begin{figure*}
    \centering
    \includegraphics[scale=0.5]{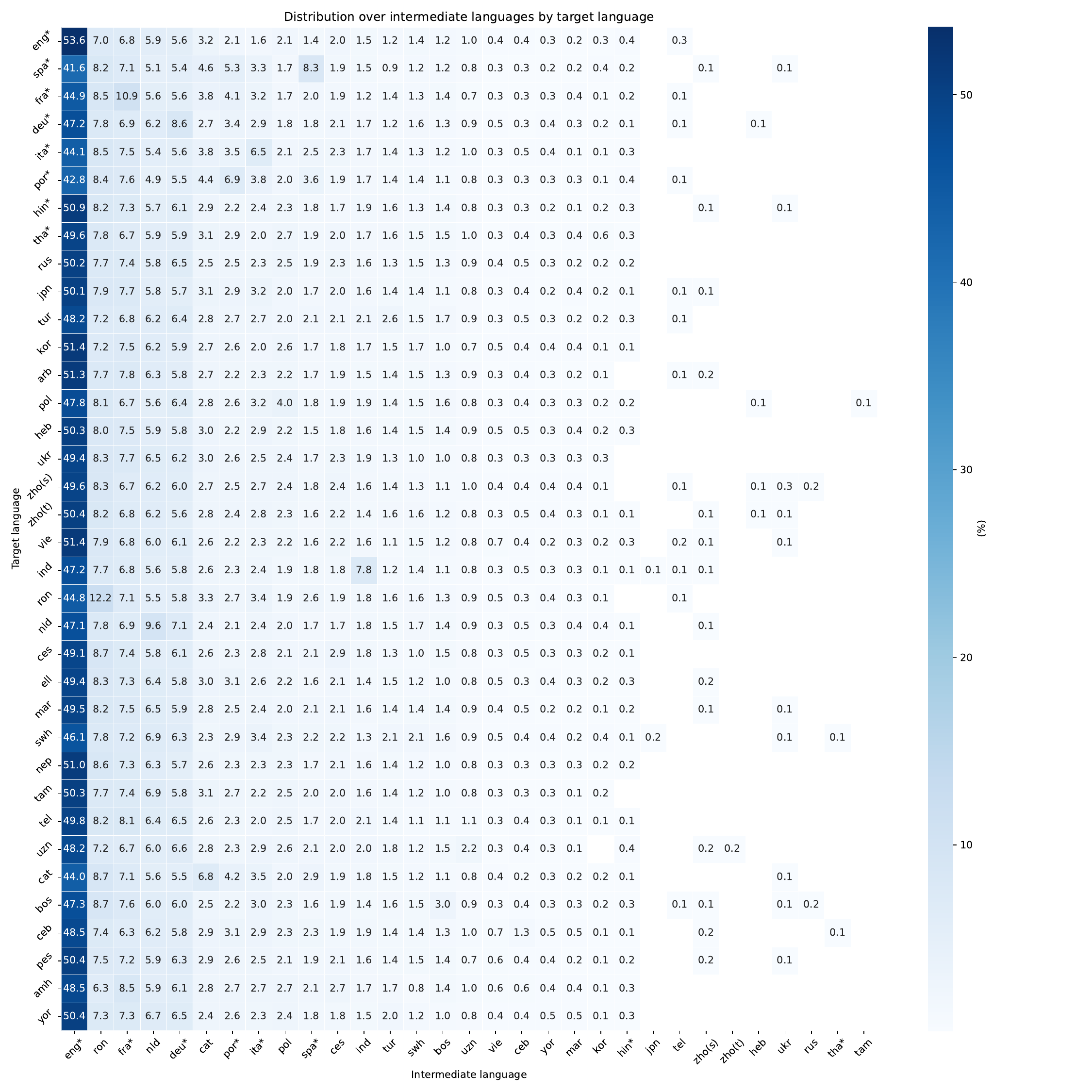}
    \caption{Distribution over intermediate languages for all correct (off-target) intermediate layer equivalents. This is an expanded version of \autoref{fig:int_lang} by target language.}
    \label{fig:int_lang_llama_by_tgt}
\end{figure*}

Given that there is a prior model-internal reasoning step before generation, we are interested in the language(s) in which this reasoning occurs.
As we visualize in \autoref{fig:layerwise_evolution} and \autoref{fig:evolution-grid}, there is a somewhat fluid shift to the target language at later layers; we can consider these the layers responsible for the translation / generation part of the pipeline. 
We see that this shift emerges around the fourth-last layer; we therefore consider the task-solving layers as all the layers up to the fourth-last layer. 
We show the distribution of correct intermediate equivalents over intermediate languages for each target language (averaged over source language) in \autoref{fig:int_lang_llama_by_tgt} for \llama{} for these middle layers.
In general, we see that this distribution looks similar over target languages, supporting the idea of model interlingua that is to a large degree agnostic of the target language.
The plot looks similar for \aya{} and also largely similar when using different layer cutoffs for what we consider as task-solving layers, with increasing target language presence as we move closer to the final layer.

\newpage

\section{On-targetness}
\label{sec:on_targetness}

\begin{figure*}
    \centering
    \includegraphics[scale=0.4]{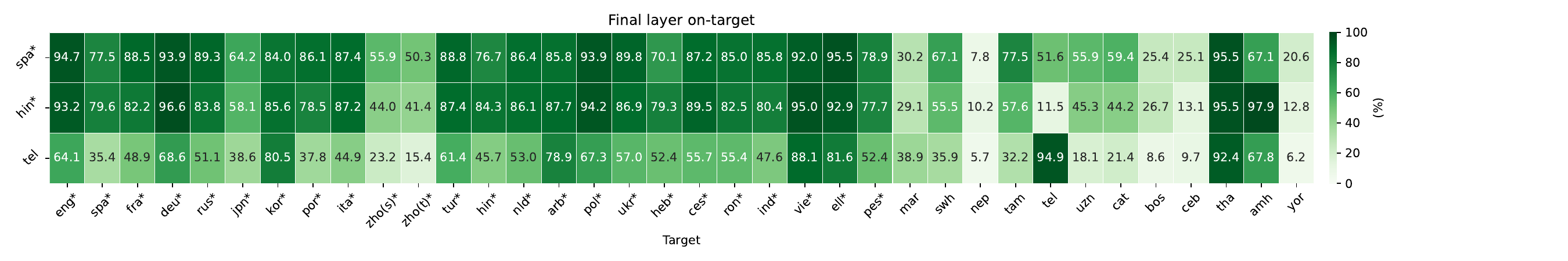}
    \includegraphics[scale=0.4]{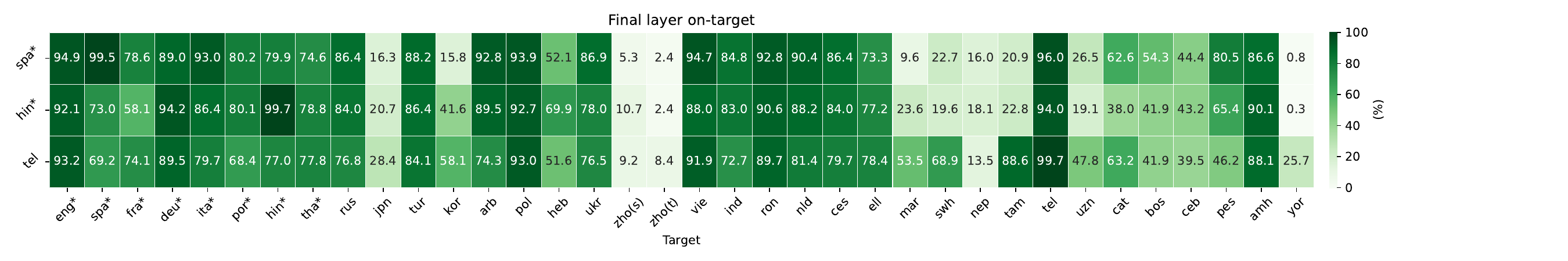}
    \caption{Percentage of outputs that were on-target for \aya{} (top) and \llama{} (bottom). * supported language}
    \label{fig:on-targetness}
\end{figure*}

We show the percentage of final layer on-target outputs in \autoref{fig:on-targetness} for both models. 
Predictably, this is higher for supported HRLs than for unsupported or low-resource languages. 
Note that these results are dependent on the accuracy of the language ID model, which may be easily confused on short spans of text.

\newpage

\section{Results for all languages}
\label{sec:all_results}

\begin{figure*}[!ht]
    \centering
    \hspace*{-10mm}\includegraphics[scale=0.7]{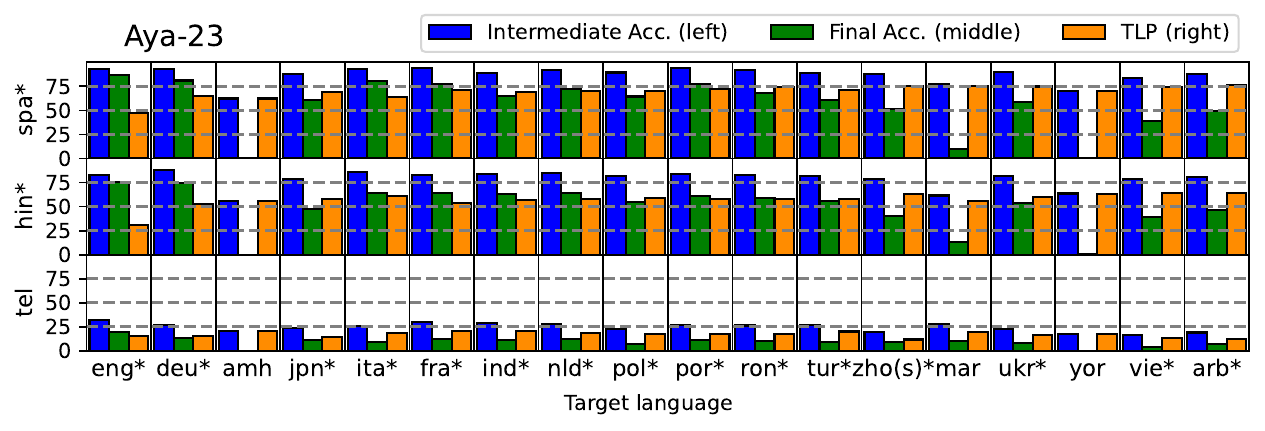}
    \hspace*{-10mm}\includegraphics[scale=0.7]{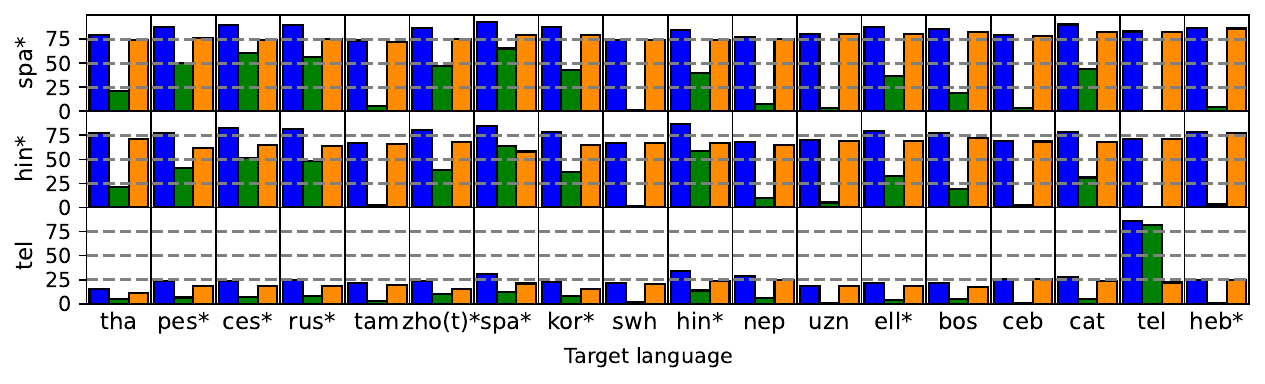}
    \caption{Intermediate and final accuracy, and \tlp{} (\autoref{eqn:tlp}) for \aya{}, sorted in ascending order of mean \tlp{} (complete version of \autoref{fig:main_results}, for all languages).}
    \label{fig:main_results_complete}
\end{figure*}

\begin{figure*}[!ht]
    \centering
    \hspace*{-10mm}\includegraphics[scale=0.7]{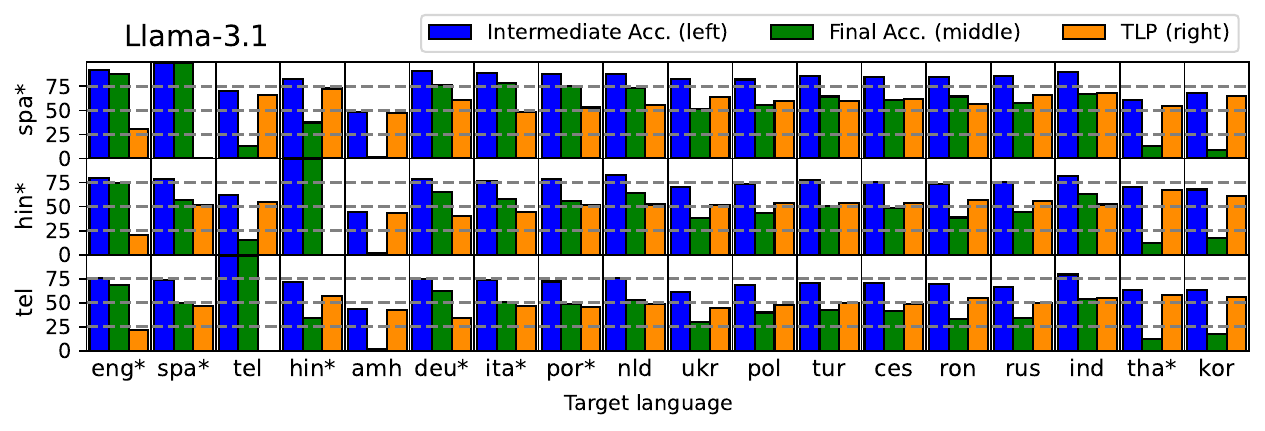}
    \hspace*{-10mm}\includegraphics[scale=0.7]{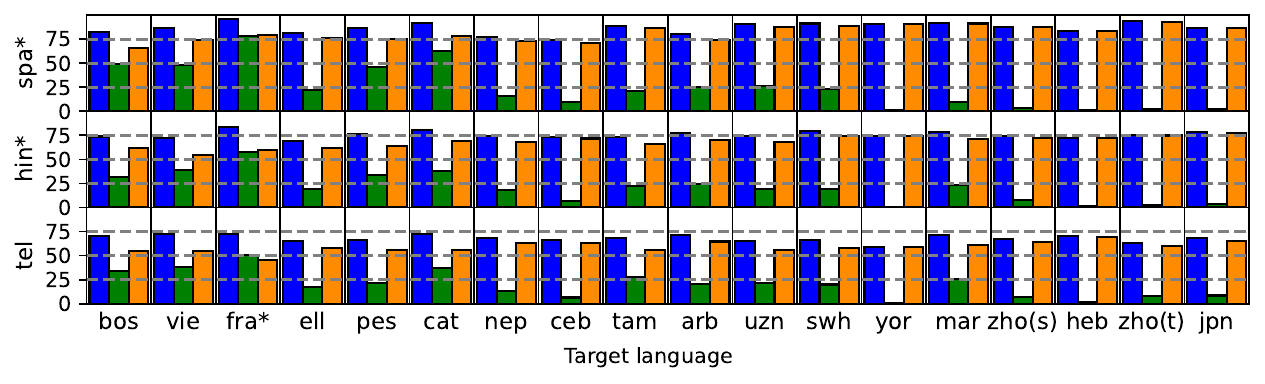}
    \caption{Intermediate and final accuracy, and \tlp{} (\autoref{eqn:tlp}) for \llama{}, sorted in ascending order of mean \tlp{} (complete version of \autoref{fig:main_results}, for all languages).}
    \label{fig:main_results_complete_llama}
\end{figure*}

We show the expanded version of \autoref{fig:main_results} in \autoref{fig:main_results_complete} for \aya{} and \autoref{fig:main_results_complete_llama} for \llama.

\newpage
\section{Arithmetic task}

\label{app:arithmetic}
See examples of our input, intermediate responses, and the final generated responses in \autoref{app:arithmetic_eg1} and \autoref{app:arithmetic_eg2}.
We tried various prompting settings, including a zero-shot prompt. We found that both \aya~ and \llama~ tended to give sentence-long responses in those settings. 
The above format, i.e. 1-shot prompting with the example in the target language, worked best to constrain the answer.

We also tried a variant of this task where the input itself in provided in the target language. Our main observations were that even in this case, the intermediate layers exhibit answers in HRLs before switching to the target language. Further, this degrades performance for all languages, especially in Marathi and Swahili. 
This ties back to our discussion in \autoref{sec:discussion} about source language comprehensibility.

\begin{table*}
    \includegraphics[scale=1]{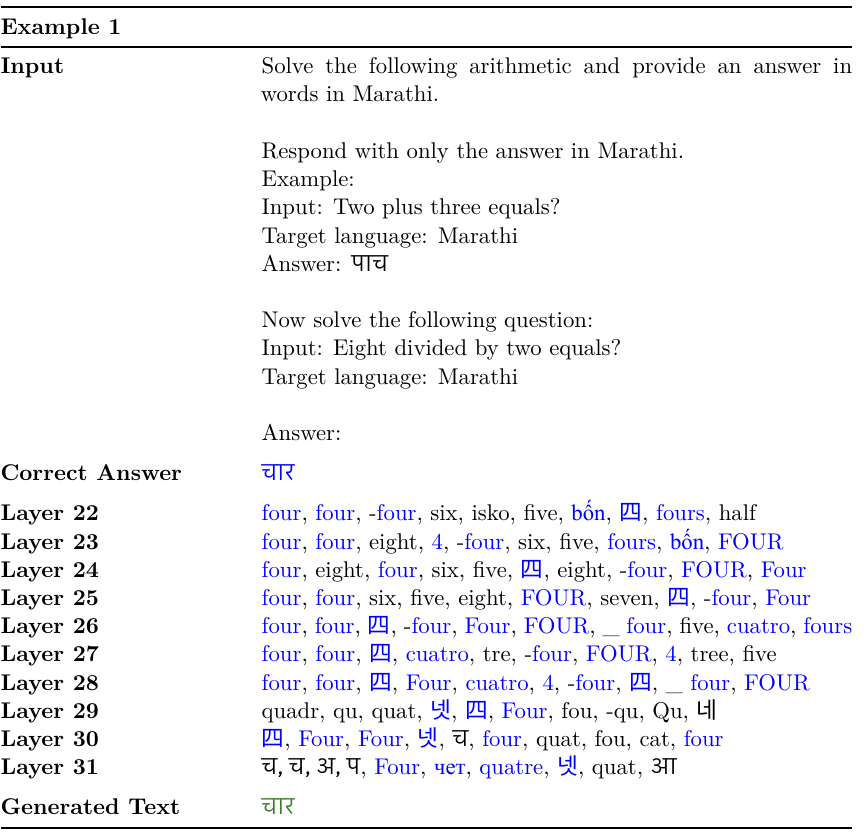}
    \caption{Input and intermediate and final response for the word arithmetic task with Marathi. Intermediate correct responses are highlighted in \textcolor{blue}{blue}.}
    \label{app:arithmetic_eg1}
\end{table*}

\begin{table*}
    \includegraphics[scale=1]{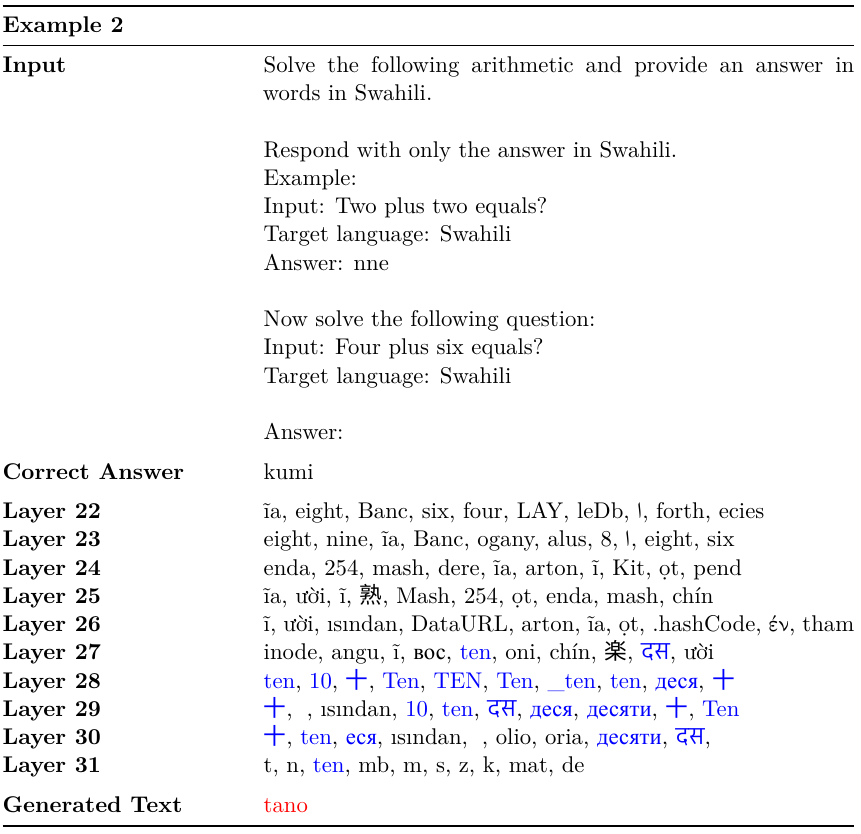}
    \caption{Input and intermediate and final response for the word arithmetic task with Marathi. Intermediate correct responses are highlighted in \textcolor{blue}{blue}. We see that even though the correct answer is discovered in intermediate layers, the final answer is wrong.}
    \label{app:arithmetic_eg2}
\end{table*}

\newpage
\section{Examples for sentence-level translation}
While this work focuses on single word translation, we show examples of applying iterative logit lens as described in \autoref{sec:method} for a sentence-level translation task. We observe that intermediate layers exhibit code-switching behavior that interpolates between the correct target language and English. We invite future work in studying the translation barrier for sequence-level tasks.

\subsection{Example 1}
\begin{tabular}{@{}p{4cm}p{10cm}@{}}
\toprule
\textbf{Source Sentence} & Mes 3 minutes n'ont pas déjà commencé, n'est-ce-pas ? \\
\addlinespace
\textbf{Correct Translation} & \begin{CJK*}{UTF8}{gbsn}我的三分钟还没开始吧？\end{CJK*} \\
\addlinespace
\textbf{Layer 32 (Final Layer)} & \colorbox{good}{\begin{CJK*}{UTF8}{gbsn}我的3分钟还没有开始，是吗？\end{CJK*}} \\
\textbf{Layer 29}                & \colorbox{partial}{\begin{CJK*}{UTF8}{gbsn}我的3 minute还没有开始，是吗？\end{CJK*}} \\
\textbf{Layer 27}                & \colorbox{bad}{my three minutes chua chua\begin{CJK*}{UTF8}{gbsn}开始吧\end{CJK*} right\begin{CJK*}{UTF8}{gbsn}吗\end{CJK*}?} \\
\textbf{Layer 25}                & \colorbox{bad}{my three minutes already chua\begin{CJK*}{UTF8}{gbsn}开始吧\end{CJK*} right\begin{CJK*}{UTF8}{gbsn}{吧}\end{CJK*}?} \\
\bottomrule
\end{tabular}

\bigskip

\subsection{Example 2}
\begin{tabular}{@{}p{4cm}p{10cm}@{}}
\toprule
\textbf{Source Sentence} & Je veux dire, je suis déjà assez nerveuse comme ça. \\
\addlinespace
\textbf{Correct Translation} & \begin{CJK*}{UTF8}{gbsn}我是说，我已经够紧张的了\end{CJK*} \\
\addlinespace
\textbf{Layer 32 (Final Layer)} & \colorbox{good}{\begin{CJK*}{UTF8}{gbsn}我的意思是，我已经很紧张了。\end{CJK*}} \\
\textbf{Layer 29}                & \colorbox{partial}{\begin{CJK*}{UTF8}{gbsn}我的意思是我我 already足够紧张了。\end{CJK*}} \\
\textbf{Layer 27}                & \colorbox{bad}{\begin{CJK*}{UTF8}{gbsn}我mean是我我 already cukup nervous了。\end{CJK*}} \\
\textbf{Layer 25}                & \colorbox{bad}{\begin{CJK*}{UTF8}{gbsn}我 already là, already already already nervous了。\end{CJK*}} \\
\bottomrule
\end{tabular}

\subsection{Example 3}
\begin{tabular}{@{}p{4cm}p{10cm}@{}}
\toprule
\textbf{Source Sentence} & De plus, au Canada, nous avons un formidable système de santé. \\
\addlinespace
\textbf{Correct Translation} & \begin{CJK*}{UTF8}{gbsn}在加拿大，我们有很好的医保制度\end{CJK*} \\
\addlinespace
\textbf{Layer 32 (Final Layer)} & \colorbox{good}{\begin{CJK*}{UTF8}{gbsn}此外，在加拿大，我们有一个很棒的医疗系统。\end{CJK*}} \\
\textbf{Layer 29}                & \colorbox{partial}{\begin{CJK*}{UTF8}{gbsn}此外，我们加拿大，我们拥有 fantast好的的健康 system。\end{CJK*}} \\
\textbf{Layer 27}                & \colorbox{bad}{\begin{CJK*}{UTF8}{gbsn}此外，我们 Canada我们我们有 fantast好的的 healthcare system。\end{CJK*}} \\
\textbf{Layer 25}                & \colorbox{bad}{\begin{CJK*}{UTF8}{gbsn}此外，我们 Canada we我们有 superb好的的 healthcare system。\end{CJK*}} \\
\bottomrule
\end{tabular}

\bigskip

\subsection{Example 3}
\begin{tabular}{@{}p{4cm}p{10cm}@{}}
\toprule
\textbf{Source Sentence} & Mais, tout d'abord, vous devez savoir 2 choses à mon propos \\
\addlinespace
\textbf{Correct Translation} & Aber Sie müssen zuerst zwei Dinge über mich wissen. \\
\addlinespace
\textbf{Layer 32 (Final Layer)} & \colorbox{good}{Aber zuerst, Sie müssen 2 Dinge über mich wissen.} \\
\textbf{Layer 29}                & \colorbox{partial}{But zuerst müssen ihr müssen zwei2 things über mich wetenissen.} \\
\textbf{Layer 27}                & \colorbox{bad}{but first einmal you must know two things things about me know know.} \\
\bottomrule
\end{tabular}

\newpage

\section{More Related Work}

\paragraph{Cross-lingual Representations}
Parallel sentences exhibit highly similar hidden states in early layers of multilingual LLMs \citep{muller-etal-2021-first, conneau-etal-2020-emerging}, and fine-tuning solely on English tasks further strengthens this alignment \citep{li-murray-2023-zero}, enabling cross-lingual transfer. Moreover, the degree of representation similarity predicts performance in other languages \citep{bayling, Li_Shi_Liu_Yang_Payani_Liu_Du_2025, liu2025middlelayerrepresentationalignmentcrosslingual, li-etal-2024-improving-context}. 
Our work adds to this discourse by characterizing the task-solving languages of the model, and providing evidence of some degree of target-language agnosticity of the task-solving stage.

%% file: latex/tables/language_table.tex
\begin{table*}[t]
\centering
\small
\resizebox{\textwidth}{!}{
\footnotesize
\begin{tabular}{llllll}
\hline
\textbf{Language} & \textbf{Language (Sub)Family} & \textbf{FLORES+ code}  & \textbf{Supported by Aya} & \textbf{Supported by Llama} & \textbf{Resource level}  \\
\hline
English & Germanic & eng\_Latn & \checkmark & \checkmark & \(\sim \)7.0M \\
Cebuano & Austronesian &ceb\_Latn & - & -& \(\sim \)6.1M \\
German & Germanic &deu\_Latn & \checkmark & \checkmark & \(\sim \)3M\\
French & Romance &fra\_Latn & \checkmark & \checkmark & 	\(\sim \)2.7M\\

Dutch & Germanic &nld\_Latn & \checkmark & - &	\(\sim \)2.2M\\
Russian & East Slavic &rus\_Cyrl & \checkmark & - & \(\sim \)2.1M\\
Spanish & Romance &spa\_Latn & \checkmark & \checkmark & \(\sim \)2M\\
Italian & Romance &ita\_Latn & \checkmark & \checkmark& \(\sim \)1.9M\\
Polish & West Slavic &pol\_Latn & \checkmark & - & \(\sim \)1.7M\\
Chinese (Simplified) & Sinitic &zho\_Hans & \checkmark & - & \(\sim \)1.5M\\
Chinese (Traditional) &Sinitic & zho\_Hant & \checkmark & - & (see above)\\
Japanese & Japonic &jpn\_Jpan & \checkmark & - & \(\sim \)1.5M\\
Ukranian & East Slavic &ukr\_Cyrl & \checkmark & - & \(\sim \)1.4M\\
Vietnamese & Austroasiatic &vie\_Latn & \checkmark & -& \(\sim \)1.3M\\
Arabic & Semitic &arb\_Arab & \checkmark & - & \(\sim \)1.3M\\

Portuguese & Romance &por\_Latn & \checkmark & \checkmark & \(\sim \)1.2M\\

Persian & Indo-Iranian &pes\_Arab & \checkmark & -&\(\sim \)1M\\
Catalan & Romance &cat\_Latn & - & -& \(\sim \)800k\\
Indonesian & Austronesian &ind\_Latn & \checkmark & -& \(\sim \)700k\\
Korean & Koreanic &kor\_Hang & \checkmark & - & \(\sim \)700k\\
Turkish & Turkic &tur\_Latn & \checkmark & - & \(\sim \)600k\\
Czech & West Slavic &ces\_Latn & \checkmark & - &\(\sim \)600k\\
Romanian & Romance &ron\_Latn & \checkmark & - &\(\sim \)500k\\

Hebrew & Semitic &heb\_Hebr & \checkmark & -& \(\sim \)400k \\
Uzbek & Turkic &uzn\_Latn & - & -&\(\sim \)300k\\
Greek & Hellenic &ell\_Grek & \checkmark& - & \(\sim \)300k\\
Tamil & Southern Dravidian &tam\_Taml & -& -&\(\sim \)200k\\
Thai & Kra-Dai &tha\_Thai & - & \checkmark &\(\sim \)200k\\
Hindi & Indic &hin\_Deva & \checkmark & \checkmark& \(\sim \)200k\\
Telugu & Southern Dravidian &tel\_Telu & - & -&\(\sim \)100k\\
Swahili & Niger-Congo &swh\_Latn & - & - &\(\sim \)100k\\
Marathi & Indic &mar\_Deva & - & - &\(\sim \)100k\\
Bosnian & South Slavic &bos\_Latn & - & - &\(\sim \)100k\\
Yoruba & Niger-Congo&yor\_Latn & - &- & \(\sim\)30k\\
Nepali & Indo-Aryan &nep\_Deva & -& - &\(\sim \)30k\\

Amharic & Semitic &amh\_Ethi& - & -&\(\sim \)15k \\

\hline

\end{tabular}
}
\centering

\caption{All studied languages listed with model support details, with FLORES+ codes from \citet{nllb-24}. We report the number of Wikipedia articles for a language as a proxy for its resource level, with the anomaly of Cebuano Wikipedia being disproportionately large due to \href{https://en.wikipedia.org/wiki/Cebuano_Wikipedia}{automatic creation}.}
\label{table:languages}
\end{table*}

%% file: latex/tables/lexica_equivalents.tex
\begin{table*}[t]

\resizebox{0.5\textwidth}{!}{
\footnotesize
\begin{tabular}{lc}
\hline
\textbf{Language} & \textbf{Average Number of Equivalents in Lexica}  \\
\hline

English & 7.5 \\

Spanish & 9.5 \\

French & 7.7 \\

German & 9.8\\

Russian &  12.9\\

Japanese & 6.0 \\

Portuguese & 8.6 \\

Italian & 7.4 \\

Turkish & 9.7 \\

Korean &  8.6\\

Arabic & 6.8 \\

Polish & 10.3\\

Hindi & 7.0\\

Hebrew & 3.9 \\

Ukranian & 7.4 \\

Chinese (Simplified) & 1 \\
Chinese (Traditional) & 1\\
Vietnamese & 6.8\\
Indonesian & 9.0 \\
Romanian & 12.9 \\
Dutch & 8.2 \\
Czech &  6.5\\
Greek & 2.7\\
Marathi & 2.3 \\
Swahili & 1.7 \\
Nepali &  1.1\\
Tamil &1.9 \\
Telugu & 1\\
Uzbek & 4.4\\
Catalan & 1.4\\
Bosnian &  5.5\\
Cebuano & 1.3\\
Persian &5.4 \\
Thai & 9.4 \\
Amharic & 1 \\
Yoruba & 1.4\\

\hline

\end{tabular}
}
\centering

\caption{All studied languages listed with average number of equivalents per source word per language. }
\label{table:lexica_equivalents}
\end{table*}